\def\year{2020}\relax
\documentclass[letterpaper]{article} 
\usepackage{aaai20}  
\usepackage{times}  
\usepackage{helvet} 
\usepackage{courier}  
\usepackage[hyphens]{url}  
\usepackage{graphicx} 
\usepackage{graphicx}  
\usepackage{amsmath}
\usepackage{amsfonts}       
\usepackage{amssymb}

\def\myproof{1} 
\title{Scalable Variational Bayesian Kernel Selection for\\ Sparse Gaussian Process Regression}
\author{Tong Teng,\textsuperscript{\rm 1}\thanks{Equal contribution.} Jie Chen,\textsuperscript{\rm 2$\ast$} Yehong Zhang,\textsuperscript{\rm 1$\ast$} Bryan Kian Hsiang Low\textsuperscript{\rm 1}\\ 
	\textsuperscript{\rm 1}Department of Computer Science, National University of Singapore, Republic of Singapore\\
	\textsuperscript{\rm 2}College of Computer Science and Software Engineering, Shenzhen University, P. R. China\\
	\{tengtong, yehong, lowkh\}@comp.nus.edu.sg, chenjie@szu.edu.cn
}
\begin{document}
	
\maketitle

\begin{abstract}
This paper presents a \emph{variational Bayesian kernel selection} (VBKS) algorithm for  \emph{sparse Gaussian process regression} (SGPR) models. In contrast to existing GP kernel selection algorithms that aim to select only one kernel with the highest model evidence, our VBKS algorithm considers the kernel as a random variable and learns its belief from data such that the uncertainty of the kernel can be interpreted and exploited to avoid overconfident GP predictions. To achieve this, we represent the probabilistic kernel as an additional variational variable in a \emph{variational inference} (VI) framework for SGPR models where its posterior belief is learned together with that of the other variational variables (i.e., inducing variables and kernel hyperparameters). In particular, we transform the discrete kernel belief into a continuous parametric distribution via reparameterization in order to apply VI. Though it is computationally challenging to jointly optimize a large number of hyperparameters due to many kernels being evaluated simultaneously by our VBKS algorithm, we show that the variational lower bound of the log-marginal likelihood can be decomposed into an additive form such that each additive term depends only on a disjoint subset of the variational variables and can thus be optimized independently. Stochastic optimization is then used to maximize the variational lower bound by iteratively improving the variational approximation of the exact posterior belief via stochastic gradient ascent, which incurs constant time per iteration and hence scales to big data. We empirically evaluate the performance of our VBKS algorithm on  synthetic and massive real-world datasets.
\end{abstract}

\section{Introduction}
\label{intro}
A \emph{Gaussian process regression} (GPR) model is a kernel-based Bayesian nonparametric model that represents the correlation/similarity of the data using a kernel for performing nonlinear probabilistic regression.
A limitation of such a \emph{full-rank} GPR model is its poor scalability to big data since computing its predictive belief and learning the kernel hyperparameters incur cubic time in the data size.
To overcome this limitation, a number of \emph{sparse GPR} (SGPR) models have been proposed \cite{LowUAI12,LowUAI13,LowTASE15,gal2015,hensman2015,Hoang2015,HoangICML16,hoang2017,lazaro2010,LowDyDESS15,LowAAAI15,Quinonero2005,Titsias2009,Titsias2013,LowAAAI14,Yu2019}. These SGPR models exploit a small set of inducing variables or a spectral representation of the kernel to derive a low-rank GP approximation for achieving scalable training and prediction to big data.

All the SGPR models mentioned above are either designed only for the widely-used \emph{squared exponential} (SE) kernel \cite{Titsias2013,Yu2019} or assume the kernel type to be specified by the user \emph{a priori}.
However, in the era of big data, 
it has become all but impossible for non-experts to 
manually select an appropriate kernel for a GP model 
since the underlying correlation structures of massive datasets are usually too complex to be captured by the commonly-used base kernels (e.g., SE and periodic kernels). Such an issue can be resolved by kernel selection algorithms \cite{duvenaud13,kim2018,lu2018,malkomes2016} which are designed to automatically find a kernel with the highest model evidence (e.g., marginal likelihood, Bayesian information criterion) given a dataset. 
These algorithms have demonstrated success in improving the GP predictive performance over that of using the manually specified kernel. But, selecting only one kernel with the highest model evidence and ignoring the uncertainty of it being the true kernel may result in overconfident predictions, especially if other kernels yield similar model evidences \cite{hoeting1999}.
This motivates the need to design and develop a \emph{Bayesian kernel selection} (BKS) algorithm that, instead of searching for the \emph{best} kernel, considers the kernel as a random variable defined over a kernel space and learns a belief of the probabilistic kernel from data such that the uncertainty of the kernel can be interpreted and exploited to avoid overconfident GP predictions, which is the focus of our work here.

Most existing BKS algorithms for GP models approximate the kernels using their spectral density representation and thus only work for stationary kernels \cite{benton2019,oliva2016,wilson2013}.
The BKS algorithm of~\citeauthor{malkomes16b}~\shortcite{malkomes16b} caters to any kernel but is designed for the \emph{full-rank} GPR model only.
So, its approach of updating the kernel belief scales poorly in the data size.
%
This paper presents a \emph{variational BKS} (VBKS) algorithm for SGPR models without any stationary assumption on the kernels.
In particular, we represent the probabilistic kernel as an additional variational variable in a \emph{variational inference} (VI) framework for the SGPR models where its posterior belief
is learned together with that of the other variational variables (i.e., inducing variables and hyperparameters) by minimizing the \emph{Kullback-Leibler} (KL) divergence between a variational approximation and their exact posterior belief or, equivalently, maximizing a variational lower bound of the log-marginal likelihood (Section~\ref{VBKS}).
Unfortunately, the 
existing variational SGPR models
\cite{hensman2015,Hoang2015,HoangICML16,Titsias2009,Titsias2013,Yu2019} cannot be straightforwardly used by our VBKS algorithm since (a) they commonly use continuous distributions (e.g., normal distribution) that cannot directly accommodate the discrete kernel belief involving further constraints, 
and (b)
the VBKS algorithm has to jointly learn the posterior belief of a large number of hyperparameters from big data as \emph{many} kernels of \emph{different} types are evaluated simultaneously, 
which is computationally more expensive than approximating the posterior belief of hyperparameters for only one specified kernel.

To address the above challenges, we first reparameterize the discrete kernel belief using variational variables to a continuous parametric distribution (Section~\ref{VBKS}) and then decompose the variational lower bound into an additive form such that each additive term depends only on a disjoint subset of the variational variables and can thus be maximized independently (Section~\ref{SBKS}).
To maximize the variational lower bound, stochastic optimization is used to iteratively improve the variational approximation of the exact posterior belief via stochastic gradient ascent where the stochastic gradient
is estimated by first sampling from the variational distributions and then from the observed data (Section~\ref{opti}). The former sampling step makes the gradient estimation applicable to any differentiable kernel function in the kernel space (rather than only the SE kernel adopted in the works of \citeauthor{Titsias2013}~\shortcite{Titsias2013} and \citeauthor{Yu2019}~\shortcite{Yu2019}) while the latter step enables our VBKS algorithm to incur constant time per iteration and hence scale to big data.
Consequently, the kernel posterior belief can be updated as and when more training data is used, which makes it possible to 
perform BKS
without using the full massive dataset and thus achieve competitive predictive performance fast.
%
We empirically evaluate the performance of our VBKS algorithm on synthetic and two massive real-world datasets.
%
%
%
\section{Background and Notations}
\label{bg}
\subsection{Gaussian Process Regression (GPR)}
Let $\mathcal{X}$ denote a $d$-dimensional input domain such that each input vector $\mathbf{x} \in \mathcal{X}$ is associated with a noisy output $y(\mathbf{x} )\triangleq f(\mathbf{x}) + \epsilon$ observed from corrupting the function $f$ evaluated at $\mathbf{x}$ by an additive noise $\epsilon \sim \mathcal{N}(0, \sigma^2_n)$ where $\sigma^2_n$ is the noise variance. Let $\{f(\mathbf{x})\}_{\mathbf{x} \in \mathcal{X}}$ denote a \emph{Gaussian process} (GP), that is, every finite subset of $\{f(\mathbf{x})\}_{\mathbf{x} \in \mathcal{X}}$ follows a multivariate Gaussian distribution.
Such a GP is fully specified by its \emph{prior} mean $\mathbb{E}[f(\mathbf{x})]$ and covariance $k(\mathbf{x}, \mathbf{x}') \triangleq \text{cov}[f(\mathbf{x}), f(\mathbf{x}')]$ for all $\mathbf{x}, \mathbf{x}' \in \mathcal{X}$, the latter of which is usually defined by one of the widely-used kernels (e.g., \emph{squared exponential} (SE), \emph{periodic} (PER)) 
with a vector of hyperparameters $\boldsymbol{\theta}_k$.
In this paper, $\mathbb{E}[f(\mathbf{x})]$ is assumed to be zero and $f(\mathbf{x}; k)$ is used to denote a function $f(\mathbf{x})$ with GP prior covariance $k(\mathbf{x}, \mathbf{x}')$ for notational simplicity.

Supposing a column vector $\mathbf{y}_\mathcal{D} \triangleq (y(\mathbf{x}))^\top_{\mathbf{x} \in \mathcal{D}}$ of noisy outputs are observed by evaluating function $f$ at a set $\mathcal{D} \subset \mathcal{X}$ of training inputs, a full-rank GPR model can perform probabilistic regression by providing a GP predictive belief $p(f(\mathbf{x}_*)|\mathbf{y}_\mathcal{D}) \triangleq \mathcal{N}(\mu_{\mathbf{x}_*|\mathcal{D}}, \sigma^2_{\mathbf{x}_*|\mathcal{D}})$ for any test input $\mathbf{x}_* \in \mathcal{X}$ with the following \emph{posterior} mean and variance:
\begin{equation}\label{GPpred}
\begin{array}{rl}
\mu_{\mathbf{x}_*|\mathcal{D}} \hspace{-2.4mm} &\triangleq \Sigma_{\mathbf{x}_* \mathcal{D}} (\Sigma_{\mathcal{D} \mathcal{D}} + \sigma^2_n I)^{-1}\mathbf{y}_\mathcal{D} \vspace{1mm}\\
\sigma^2_{\mathbf{x}_*|\mathcal{D}} \hspace{-2.4mm} &\triangleq k(\mathbf{x}_*, \mathbf{x}_*) - \Sigma_{\mathbf{x}_* \mathcal{D}} (\Sigma_{\mathcal{D} \mathcal{D}} + \sigma^2_n I)^{-1} \Sigma_{ \mathcal{D}\mathbf{x}_*}
\end{array}
\end{equation}
where $\Sigma_{\mathbf{x}_* \mathcal{D}} \triangleq (k(\mathbf{x}_*, \mathbf{x}))_{\mathbf{x} \in \mathcal{D}}$, $\Sigma_{\mathcal{D} \mathcal{D}} \triangleq (k(\mathbf{x}, \mathbf{x}'))_{\mathbf{x}, \mathbf{x}' \in \mathcal{D}}$, and $\Sigma_{ \mathcal{D}\mathbf{x}_*} \triangleq \Sigma^\top_{\mathbf{x}_* \mathcal{D}}$.
Due to the inversion of $\Sigma_{\mathcal{D} \mathcal{D}} + \sigma^2_n I$, computing the above predictive belief incurs $\mathcal{O}(|\mathcal{D}|^3)$ time and thus scales poorly in the size $|\mathcal{D}|$ of observed data.
%
\subsection{Sparse Gaussian Process Regression (SGPR)}
To improve the scalability of the GP model, a number of SGPR models have been proposed. 
These SGPR models exploit a vector $\mathbf{u}_k \triangleq \{f(\mathbf{x}; k)\}_{\mathbf{x} \in \mathcal{U}}$ of inducing variables\footnote{Let $\mathbf{u}_k$ denote a vector of inducing variables whose prior covariance is computed using the kernel function $k(\mathbf{x}, \mathbf{x}')$.} for a small set $\mathcal{U} \subset \mathcal{X}$ of inducing inputs (i.e., $|\mathcal{U}| \ll |\mathcal{
D}|$) for approximating the GP predictive belief:
\begin{equation}\label{pred}
\begin{array}{rl}
p(f(\mathbf{x}_*)|\mathbf{y}_\mathcal{D}) \hspace{-2.4mm}&= \displaystyle \int p(f(\mathbf{x}_*) | \boldsymbol{\alpha}, \mathbf{y}_\mathcal{D}) \ p(\boldsymbol{\alpha} | \mathbf{y}_\mathcal{D}) \ \text{d} \boldsymbol{\alpha} \\
\hspace{-2.4mm}&\approx \displaystyle \int q(f(\mathbf{x}_*) | \boldsymbol{\alpha}, \mathbf{y}_\mathcal{D}) \  q(\boldsymbol{\alpha}) \ \text{d}\boldsymbol{\alpha}
\end{array}
\end{equation}
where $\boldsymbol{\alpha}$ is a vector of variables that can be set as either $\boldsymbol{\alpha} \triangleq \mathbf{u}_k$ (i.e., $\boldsymbol{\theta}_k$ is assumed to be a point estimate) \cite{Hoang2015,HoangICML16,Quinonero2005,Titsias2009} or $\boldsymbol{\alpha} \triangleq \text{vec}(\mathbf{u}_k, \boldsymbol{\theta}_k)$ \cite{hensman2015,Titsias2013,Yu2019}.
%
Variational inference has been used to derive $q(\boldsymbol{\alpha})$ by minimizing the KL divergence between $q(\boldsymbol{\alpha})$ and the exact posterior belief $p(\boldsymbol{\alpha} | \mathbf{y}_\mathcal{D})$.
Various conditional independence assumptions of $f(\mathbf{x}_*)$ and $\mathbf{y}_\mathcal{D}$ given $\boldsymbol{\alpha}$ have been imposed for computing $q(f(\mathbf{x}_*) | \boldsymbol{\alpha}, \mathbf{y}_\mathcal{D})$, which result in different sparse GP approximations \cite{Hoang2015,HoangICML16,Quinonero2005}. 
\subsection{Automatic Kernel Selection}\label{KS}
All the GPR and SGPR models mentioned above assume the kernel type $k(\mathbf{x}, \mathbf{x}')$ to be specified by the user and learn $\mathbf{u}_k$ and $\boldsymbol{\theta}_k$ from the data. However, the kernel choice is critical to the performance of the (sparse) GP models since various kernel types can capture different underlying correlation structures of the data (see Chapter $4$ in~\cite{Rasmussen2006} for a detailed discussion of various kernels).
Let $\mathcal{K}$ be a set of candidate kernels (e.g., SE, PER). The automatic kernel selection algorithms \cite{duvenaud13,kim2018,lloyd2014,lu2018,malkomes2016} aim to automatically find a kernel  $k \in \mathcal{K}$ with the highest model evidence (e.g., marginal likelihood, Bayesian information criterion).
%
Since the sum or product of two valid kernels (i.e., positive semidefinite kernels that define valid covariance functions) is still a valid covariance function, the kernel space $\mathcal{K}$ can be constructed by repeatedly applying the following composition rules:
$$
\begin{array}{rl}
k_3 (\mathbf{x}, \mathbf{x}^\prime) \hspace{-2.5mm}&= k_1(\mathbf{x}, \mathbf{x}^\prime) + k_2(\mathbf{x}, \mathbf{x}^\prime) \vspace{1mm}\\
k_4 (\mathbf{x}, \mathbf{x}^\prime) \hspace{-2.5mm}&= k_1(\mathbf{x}, \mathbf{x}^\prime) \times k_2(\mathbf{x}, \mathbf{x}^\prime)
\end{array}
$$
where $k_1$ and $k_2$ can be either one of the base kernels (i.e., SE, PER, \emph{linear} (LIN), and \emph{rational-quadratic} (RQ)) or a composite kernel \cite{duvenaud13}.
%
%
\section{Variational Bayesian Kernel Selection (VBKS) for SGPR Models}
\label{VBKS}
As mentioned in Section~\ref{intro}, most existing kernel selection algorithms aim to find only \emph{one} kernel $k \in \mathcal{K}$ with the highest model evidence \cite{duvenaud13,kim2018,lloyd2014,lu2018,malkomes2016}. However, if several kernels achieve similar model evidences, ignoring the uncertainty among the kernels and selecting only one kernel with the highest model evidence may result in overconfident inferences/predictions or overfitting, especially if some composite kernel structures in the kernel space are complex.
To resolve this issue, the \emph{Bayesian kernel selection} (BKS) problem considers $k$ as a random variable and introduces a kernel belief $p(k)$ over $k$. Then, the GP predictive belief~\eqref{pred} has to consider an additional variable $k$ of the kernel, which yields
\begin{equation}\label{predk}
p(f(\mathbf{x}_*)|\mathbf{y}_\mathcal{D}) = \mathbb{E}_{p( \mathbf{f}_\mathcal{D}, \boldsymbol{\alpha}, k | \mathbf{y}_\mathcal{D})} [p(f(\mathbf{x}_*) | \mathbf{f}_\mathcal{D}, \boldsymbol{\alpha}, k)]
\end{equation}
where $\boldsymbol{\alpha} \triangleq \text{vec}(\mathbf{u}_k, \boldsymbol{\theta}_k)$, $\mathbf{f}_\mathcal{D} \triangleq (f(\mathbf{x}))^\top_{\mathbf{x} \in \mathcal{D}}$, and $\mathbf{u}_k$ and $\boldsymbol{\theta}_k$ still, respectively, denote the vectors of inducing variables and hyperparameters of kernel $k$ to ease notations, even though $k$ is now probabilistic. The exact definitions of $\mathbf{u}_k$, $\boldsymbol{\theta}_k$, and $p(k)$ will be introduced later. Next, the key issue is to approximate the intractable posterior belief $p( \mathbf{f}_\mathcal{D}, \boldsymbol{\alpha}, k | \mathbf{y}_\mathcal{D})$ in \eqref{predk} such that the GP predictive belief $p(f(\mathbf{x}_*)|\mathbf{y}_\mathcal{D})$ can be computed analytically and efficiently.
To achieve this, the active structure discovery algorithm of~\citeauthor{malkomes16b}~\shortcite{malkomes16b}
has proposed to approximate $p(\boldsymbol{\theta}_k|\mathbf{y}_\mathcal{D}, k)$ and $p(\mathbf{y}_\mathcal{D}|k)$ via Laplace approximation such that the posterior belief of the kernel $p(k|\mathbf{y}_\mathcal{D})$ can be computed by applying Bayes' rule. However, such a BKS algorithm is designed for the full-rank GPR model only. So, it still incurs $\mathcal{O}(|\mathcal{D}|^3)$ time and scales poorly in the size $|\mathcal{D}|$ of observed data.

To scale BKS of GP models to big data, we propose to approximate the posterior belief $p( \mathbf{f}_\mathcal{D}, \boldsymbol{\alpha}, k | \mathbf{y}_\mathcal{D})$ in \eqref{predk} via \emph{variational inference} (VI)
for SGPR models, which we call \emph{variational BKS} (VBKS)\footnote{Though our proposed VBKS algorithm performs Bayesian kernel inference instead of ``selecting" a specific kernel(s), we call it ``kernel selection'' to be consistent with the \emph{Bayesian model selection} framework \cite{Rasmussen2006}.}.
%
%
%
In particular, the intractable posterior belief $p(\mathbf{f}_\mathcal{D}, \boldsymbol{\alpha}, k | \mathbf{y}_\mathcal{D})$ in \eqref{predk} can be approximated with a variational distribution:
\begin{equation}\label{q}
q(\mathbf{f}_\mathcal{D}, \boldsymbol{\alpha}, k) \triangleq p(\mathbf{f}_\mathcal{D} | \boldsymbol{\alpha}, k)\ q(\boldsymbol{\alpha})\ q(k)
\end{equation}
by minimizing the KL divergence 
between $q(\mathbf{f}_\mathcal{D}, \boldsymbol{\alpha}, k)$ and the exact posterior belief $p( \mathbf{f}_\mathcal{D}, \boldsymbol{\alpha}, k | \mathbf{y}_\mathcal{D})$. The variational approximation in \eqref{q} is similar to those used by existing VI frameworks of SGPR models \cite{Titsias2013,Yu2019} except that an additional term $q(k)$ is included due to the probabilistic kernel $k$ whose posterior belief needs to be learned from data together with that of $\boldsymbol{\alpha}$. We will consider a finite kernel space $\mathcal{K}$ and hence a discrete distribution over $k$ in our work here.

Let $\mathcal{K} \triangleq \{k_i\}_{i=1}^K$ be a set of kernels where each $k_i$ represents a kernel  which can either be the base kernel or the composite kernel, as described in Section~\ref{KS}. The kernel belief $p(k)$ can be defined as a vector $\mathbf{p} \triangleq (p_i)_{i=1}^K$ where $p_i \triangleq p(k = k_i)$ for $i = 1, \ldots, K$.
Similarly, the variational distribution $q(k)$ can be represented by a vector $\mathbf{q} \triangleq (q_i)_{i=1}^K$ of variational parameters where $q_i \triangleq q(k = k_i)$ for $i = 1, \ldots, K$. Then, the objective of VBKS is to minimize $\text{KL}[q(\mathbf{f}_\mathcal{D}, \boldsymbol{\alpha}, k) \| p(\mathbf{f}_\mathcal{D}, \boldsymbol{\alpha}, k | \mathbf{y}_\mathcal{D})]$ w.r.t. $\mathbf{q}$ and the variational parameters of $q(\boldsymbol{\alpha})$ with the following constraints:
$$
\begin{array}{rl}
0 \leq q_i \leq 1 \ \text{for} \ i = 1, \ldots, K, \ \text{and} \ \sum_{i=1}^K q_i = 1 \ .
\end{array}
$$
A commonly-used method to solve such a constrained optimization problem is to convert it to that of unconstrained optimization via some tricks (e.g., substitution, Lagrange multiplier). In this work, we achieve this by introducing a $K$-dimensional vector of continuous variables $\mathbf{g} \in \mathbb{R}^K$ and reparameterizing $p_i$ using  
$$
\begin{array}{rl}
p(k_i | \mathbf{g}) \triangleq \exp (g_i) / \sum_{j=1}^{K} \exp (g_j)
\end{array}
$$
where $g_i$ is the $i$-th element of $\mathbf{g}$. Then, let the variational distribution $q(k) \triangleq p(k|\mathbf{g})\ q(\mathbf{g})$.
The above-mentioned constrained optimization problem is transformed to that of minimizing $\text{KL}[q(\mathbf{f}_\mathcal{D}, \boldsymbol{\alpha}, k, \mathbf{g}) \| p(\mathbf{f}_\mathcal{D}, \boldsymbol{\alpha}, k, \mathbf{g} | \mathbf{y}_\mathcal{D})]$ w.r.t. the variational parameters (i.e., detailed later) of $q(\boldsymbol{\alpha})$ and $q(\mathbf{g})$ without any constraints.
An additional benefit of introducing $\mathbf{g}$ as a vector of variational variables is that we can then place parametric multivariate distributions on $p(\mathbf{g})$ and $q(\mathbf{g})$ such that the prior knowledge of the kernel set can be included in $p(\mathbf{g})$ and the true correlations between different kernels can be learned from data by learning the covariance parameters of $q(\mathbf{g})$, which is useful in interpreting the relationship between kernels (e.g., a high correlation between $g_i$ and $g_j$ can be interpreted as a high similarity between $k_i$ and $k_j$ with potentially similar learning performances).

%
Then, minimizing $\text{KL}[q(\mathbf{f}_\mathcal{D}, \boldsymbol{\alpha}, k, \mathbf{g}) \| p(\mathbf{f}_\mathcal{D}, \boldsymbol{\alpha}, k, \mathbf{g} | \mathbf{y}_\mathcal{D})]$
is equivalent to maximizing a variational \emph{evidence lower bound} (ELBO):
\begin{equation}\label{elbo}
\begin{array}{rl}
\mathcal{L}(q)  \triangleq &\hspace{-2.4mm} \mathbb{E}_{q(\mathbf{f}_\mathcal{D}, \boldsymbol{\alpha}, k, \mathbf{g}) }
\left[ \log p(\mathbf{y}_\mathcal{D}\mid\mathbf{f}_\mathcal{D})\right] \vspace{1mm}\\
&\hspace{-2.4mm}-\text{KL}\left[q(\mathbf{f}_\mathcal{D}, \boldsymbol{\alpha}, k, \mathbf{g}) \| p(\mathbf{f}_\mathcal{D}, \boldsymbol{\alpha}, k, \mathbf{g})\right]
\end{array}
\end{equation}
since the log marginal likelihood
$
\log p(\mathbf{y}_D) = \mathcal{L}(q) + \text{KL}[q(\mathbf{f}_\mathcal{D}, \boldsymbol{\alpha}, k, \mathbf{g}) \| p(\mathbf{f}_\mathcal{D}, \boldsymbol{\alpha}, k, \mathbf{g} | \mathbf{y}_\mathcal{D})]
$
is a constant. The derivation of \eqref{elbo} is in\if\myproof1 Appendix \ref{a.elbo}.  \fi\if\myproof0 \cite{teng2020full}. \fi Unfortunately, the evaluation of $\mathcal{L}(q)$ is intractable since it contains the inverse of 
the prior covariance matrix of the inducing variables $\mathbf{u}_k$ which depends on $\boldsymbol{\theta}_k$ but cannot be analytically integrated over $\boldsymbol{\theta}_k$. Some works \cite{Titsias2013,Yu2019} have resolved this issue by introducing a standardized kernel and reparameterizing the prior covariance matrix such that its inversion does not depend on the hyperparameters. However, such a reparameterization trick cannot be applied to our work here since the standardized kernel is only defined for the SE kernel and cannot be generalized to cater to the other kernels, especially the composite ones.
The doubly stochastic VI framework of~\citeauthor{titsias2014}~\shortcite{titsias2014} has proposed to generalize the optimization of $\mathcal{L}(q)$ to any differentiable kernel function by sampling from the variational distribution, which cannot be straightforwardly used to optimize $\mathcal{L}(q)$ in VBKS  since it is designed for GP models with only one kernel and does not consider the scalability in $K$ when \emph{many} kernels have to be evaluated simultaneously.
Next, a scalable stochastic optimization method for VBKS will be introduced to circumvent the above-mentioned issues.
\section{Scalable Stochastic Optimization for VBKS}\label{SBKS}
\begin{figure}
	\centering
	\includegraphics[scale=0.9]{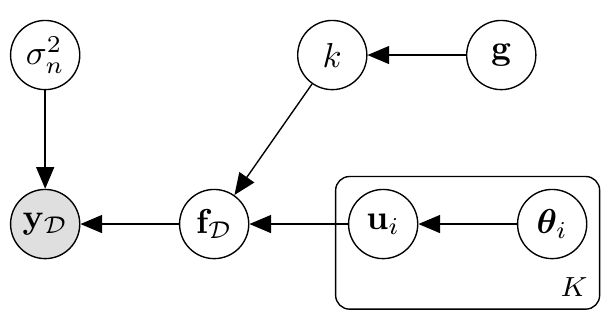}
	\caption{Graphical model of our SGPR model with the probabilistic kernel.}
	\label{fig.PGM}
\end{figure}

%
%
Let $\boldsymbol{\theta}_k \triangleq (\boldsymbol{\theta}_i)_{i = 1}^K$ and $\mathbf{u}_k \triangleq (\mathbf{u}_i)_{i=1}^K$ where $\boldsymbol{\theta}_i$ are the hyperparameters of kernel $k_i$ and $\mathbf{u}_i \triangleq (f(\mathbf{x}; k_i))_{\mathbf{x} \in \mathcal{U}}$ is a vector of inducing variables whose prior covariance is computed using kernel $k_i$.
Optimizing $\mathcal{L}(q)$ jointly w.r.t. $\boldsymbol{\theta}_k$ and $\mathbf{u}_k$ is computationally challenging since they are both high-dimensional, especially if $K$ is large. To improve the scalability of the optimization in $K$,
we assume that (a) 
$\boldsymbol{\alpha}_i \triangleq \text{vec}(\mathbf{u}_i, \boldsymbol{\theta}_i)$ for $i = 1, \ldots, K$ are independent and also independent of $k$ and $\mathbf{g}$,  
and (b) $\mathbf{f}_\mathcal{D}$ is conditionally independent of $\mathbf{g}$ given $k$. Then,
\begin{equation}
\begin{array}{l}
\hspace{-1mm} p(\mathbf{f}_\mathcal{D}, \boldsymbol{\alpha}, k, \mathbf{g}) = p(\mathbf{f}_\mathcal{D} | \boldsymbol{\alpha}, k)\ p(k|\mathbf{g})\ p(\mathbf{g}) \prod_{i=1}^K p(\boldsymbol{\alpha}_i). 
\end{array}\hspace{-2mm}
\label{p}
\end{equation}
The graphical model in Fig.~\ref{fig.PGM} shows the relationship between the variables of our SGPR model with the probabilistic kernel. 
Let the variational distribution be factorized as
\begin{equation}
\begin{array}{rl}
q(\mathbf{f}_\mathcal{D}, \boldsymbol{\alpha}, k, \mathbf{g}) = p(\mathbf{f}_\mathcal{D} | \boldsymbol{\alpha}, k)\ p(k|\mathbf{g})\ q(\mathbf{g}) \prod_{i=1}^K q(\boldsymbol{\alpha}_i)
\hspace{-1.8mm} 
\label{q1}
\end{array}
\end{equation}
where $p(\mathbf{f}_\mathcal{D} | \boldsymbol{\alpha}, k)$ and $p(k|\mathbf{g})$ are the exact posterior beliefs of $\mathbf{f}_\mathcal{D}$ and $k$.
The ELBO $\mathcal{L}(q)$ in \eqref{elbo} can be decomposed as
\begin{equation}\label{elbog}
\mathcal{L}(q) = \mathbb{E}_{q(\mathbf{g})}\left[\sum_{i=1}^{K} p(k_i | \mathbf{g})\ \mathcal{L}_{i}(q) \right]-\text{KL}\left[q(\mathbf{g})\|p(\mathbf{g})\right]
\end{equation}
where
\begin{equation}\label{Lqi}
\mathcal{L}_i (q) \hspace{-0.5mm} \triangleq \hspace{-0.5mm} \mathbb{E}_{
p(\mathbf{f}_\mathcal{D} | \boldsymbol{\alpha}_i, k_i) q(\boldsymbol{\alpha}_i)} 
[\log p(\mathbf{y}_\mathcal{D} | \mathbf{f}_\mathcal{D})]
- \text{KL}[q(\boldsymbol{\alpha}_i)||p(\boldsymbol{\alpha}_i)]
\end{equation}
includes only the variational variables $\boldsymbol{\alpha}_i$ and is thus called a \emph{local ELBO}. 
The derivation of \eqref{elbog} is in\if\myproof1 Appendix~\ref{a.elbog}. \fi\if\myproof0 \cite{teng2020full}. \fi
Interestingly, to maximize \eqref{elbog}, we can maximize each local ELBO $\mathcal{L}_i(q)$ over
$q(\boldsymbol{\alpha}_i)$ independently for $i = 1, ..., K$ and then maximize \eqref{elbog} over
$q(\mathbf{g})$ since $p(k_i|\mathbf{g}) \geq 0$ and $\mathcal{L}_i(q)$ is independent of $\mathbf{g}$ and the variational variables of $\mathcal{L}_j(q)$ for $j \neq i$, which makes the optimization of $\mathcal{L}(q)$ incur linear time in the kernel size $K$ and thus easily parallelizable.
Next, we will discuss how to maximize $\mathcal{L}(q)$ and each $\mathcal{L}_i(q)$ via stochastic gradient ascent,
which incurs only constant time per iteration and hence makes our VBKS algorithm scale well to big data.

%

\subsection{Stochastic Optimization}\label{opti}

In this section, we will first discuss the optimization of the local ELBOs $\mathcal{L}_i(q)$ for $i = 1, \ldots, K$ and then the optimization of $\mathcal{L}(q)$ via \emph{stochastic gradient ascent} (SGA).
%
\subsubsection{Optimizing the local ELBOs.}
To optimize $\mathcal{L}_i(q)$ via stochastic optimization, we reparameterize the variational variables $\mathbf{u}_i$ and $\boldsymbol{\theta}_i$ by assuming that they are affine transformations of a vector of variables which follow a standard distribution. In particular, let
$\mathbf{u}_i \triangleq \mathbf{C}_{\mathbf{u}_i} \boldsymbol{\eta}_{\mathbf{u}_i} + \mathbf{m}_{\mathbf{u}_i}$
and
$\boldsymbol{\theta}_i \triangleq \mathbf{C}_{\boldsymbol{\theta}_i} \boldsymbol{\eta}_{\boldsymbol{\theta}_i} + \mathbf{m}_{\boldsymbol{\theta}_i}$
where $\Phi_i \triangleq \text{vec}(\mathbf{C}_{\mathbf{u}_i}, \mathbf{m}_{\mathbf{u}_i}, \mathbf{C}_{\boldsymbol{\theta}_i}, \mathbf{m}_{\boldsymbol{\theta}_i})$ are variational parameters that are independent of $\mathbf{u}_i$ and $\boldsymbol{\theta}_i$, and $\boldsymbol{\mathbf{\eta}}_i \triangleq \text{vec} (\boldsymbol{\eta}_{\mathbf{u}_i}, \boldsymbol{\eta}_{\boldsymbol{\theta}_i})$ follows a standard multivariate Gaussian distribution\footnote{We use the widely-used standard multivariate Gaussian distribution as an example here. Similar to doubly stochastic VI~\cite{titsias2014}, $q(\boldsymbol{\eta}_i)$ can be any standard continuous density function which yields a different $q(\boldsymbol{\alpha}_i)$.}: $q(\boldsymbol{\mathbf{\eta}}_i) \approx p(\boldsymbol{\mathbf{\eta}}_i | \mathbf{y}_\mathcal{D}) \triangleq \mathcal{N}(\boldsymbol{\mathbf{\eta}}_i| \mathbf{0}, I)$.
We can factorize the variational distribution $q(\boldsymbol{\alpha}_i) \triangleq q(\mathbf{u}_i, \boldsymbol{\theta}_i) = q(\mathbf{u}_i)\ q(\boldsymbol{\theta}_i)$ and obtain
\begin{equation}\label{Lqi2}
q(\boldsymbol{\alpha}_i) = \mathcal{N}(\mathbf{u}_i|\mathbf{m}_{\mathbf{u}_i},  \Sigma_{\mathbf{u}_i})\ \mathcal{N}(\boldsymbol{\theta}_i|\mathbf{m}_{\boldsymbol{\theta}_i}, \Sigma_{\boldsymbol{\theta}_i})
\end{equation}
where $\Sigma_{\mathbf{u}_i} \triangleq \mathbf{C}_{\mathbf{u}_i} \mathbf{C}^\top_{\mathbf{u}_i}$ and $\Sigma_{\boldsymbol{\theta}_i} \triangleq \mathbf{C}_{\boldsymbol{\theta}_i}\mathbf{C}^\top_{\boldsymbol{\theta}_i}$.
Then, the gradient of the local ELBOs can be computed with respect to the variational parameters $\Phi_i$ using
\begin{equation}\label{grad}
\begin{array}{rl}
\nabla_{\Phi_{i}} \mathcal{L}_i  =&\hspace{-2.4mm} \mathbb{E}_{q(\boldsymbol{\eta}_i)}
\left[ \nabla_{\Phi_{i}}
\mathbb{E}_{ p(\mathbf{f}_\mathcal{D} | \boldsymbol{\alpha}_i, k_i) } [\log p(\mathbf{y}_\mathcal{D} | \mathbf{f}_\mathcal{D})]
\right]	 \vspace{1mm} \\
 &\hspace{-2.4mm}- \nabla_{\Phi_{i}} \text{KL}[q(\boldsymbol{\alpha}_i)||p(\boldsymbol{\alpha}_i)]
\end{array}
\end{equation}
The derivation of \eqref{grad} can be obtained by applying the results in Appendix D of \citeauthor{hoang2017}~\shortcite{hoang2017}.
Note that the reparameterization trick introduced above is used to make the first expectation operator $\mathbb{E}_{q(\boldsymbol{\eta}_i)}$ in \eqref{grad} independent of the variational parameters $\Phi_i$ such that the gradient operator can be moved inside the first expectation w.r.t.~$q(\boldsymbol{\eta}_i)$.
Otherwise, if we assume \eqref{Lqi2} and optimize each $\mathcal{L}_i$ in \eqref{Lqi} directly w.r.t. the variational parameters $\Phi'_i \triangleq \text{vec}(\mathbf{m}_{\mathbf{u}_i},  \Sigma_{\mathbf{u}_i}, \mathbf{m}_{\boldsymbol{\theta}_i}, \Sigma_{\boldsymbol{\theta}_i})$, then the gradient operator cannot be moved into the first expectation in \eqref{Lqi} since $\mathbb{E}_{q(\boldsymbol{\alpha}_i)}$ depends on $\Phi'_i$, which makes the estimation of the gradient $\nabla_{\Phi'_{i}}\mathcal{L}_i$ non-trivial.

Given \eqref{grad}, the gradient of $\mathcal{L}_i$ w.r.t. $\Phi_i$ can be approximated by sampling $\boldsymbol{\eta}_i \sim q(\boldsymbol{\eta}_i)$, as detailed in\if\myproof1 Appendix~\ref{a.Li}. \fi\if\myproof0 \cite{teng2020full}. \fi
%
Unfortunately, the approximation of \eqref{grad} is still computationally expensive for massive (e.g., million-sized) datasets since the estimation of $\nabla_{\Phi_{i}} \mathbb{E}_{ p(\mathbf{f}_\mathcal{D} | \boldsymbol{\alpha}_i, k_i) } [\log p(\mathbf{y}_\mathcal{D} | \mathbf{f}_\mathcal{D})]$ incurs linear time in the data size $|\mathcal{D}|$ per SGA update. To resolve this issue, we exploit the \emph{deterministic training conditional} (DTC) assumption of conditional independence among $f(\mathbf{x})$ for $\mathbf{x} \in \mathcal{D}$ given the inducing variables for deriving
\begin{equation}\label{ell}
\begin{array}{l}
\nabla_{\Phi_{i}} \mathbb{E}_{ p(\mathbf{f}_\mathcal{D} | \boldsymbol{\alpha}_i, k_i) } [\log p(\mathbf{y}_\mathcal{D} | \mathbf{f}_\mathcal{D})] \vspace{1mm}\\
= \nabla_{\Phi_{i}} \mathbb{E}_{ p(\mathbf{f}_\mathcal{D} | \boldsymbol{\alpha}_i, k_i) } [\sum_{\mathbf{x} \in \mathcal{D}}\log p(y(\mathbf{x}) | f(\mathbf{x}))] \vspace{1mm}\\
=\sum_{\mathbf{x} \in \mathcal{D}} \nabla_{\Phi_{i}} \mathbb{E}_{p(f(\mathbf{x})|\boldsymbol{\alpha}_{i}, k_i)}\left[\log p(y(\mathbf{x}) | f(\mathbf{x}))\right] \ .
\end{array}
\end{equation}
Then, we can obtain an unbiased stochastic gradient estimate of $\mathcal{L}_i$ w.r.t. $\Phi_{i}$ by uniformly sampling a \emph{mini-batch} $\mathbf{y}_{\widetilde{\mathcal{D}}}$ of the observed data where $\widetilde{\mathcal{D}}\subseteq\mathcal{D}$ and the (fixed) batch size $|\widetilde{\mathcal{D}}|$
is much smaller than $|\mathcal{D}|$. 
As a result, the SGA update of $\Phi_i$ incurs only constant time per iteration.
%
%
\subsubsection{Optimizing $\mathcal{L}(q)$ w.r.t. $q(\mathbf{g})$.}
The optimization of $\mathcal{L}(q)$ can be performed by applying similar reparameterization trick and SGA. Specifically,
let $\mathcal{L}^*_i$ be the maximal local ELBO achieved using the above SGA and $\mathbf{g} \triangleq \mathbf{C}_\mathbf{g} \boldsymbol{\eta}_\mathbf{g} + \mathbf{m}_\mathbf{g}$ with $q(\boldsymbol{\eta}_\mathbf{g}) \triangleq \mathcal{N}(\boldsymbol{\eta}_\mathbf{g}| \mathbf{0}, I)$.
The gradient of $\mathcal{L}(q)$ can be derived in the same manner as that of~\eqref{grad}:
$$
\nabla_{\Phi_\mathbf{g}} \mathcal{L} \hspace{-0.5mm} = \hspace{-0.5mm} \mathbb{E}_{q(\boldsymbol{\eta}_\mathbf{g})}\hspace{-1.5mm}\left[\sum_{i=1}^{K} \mathcal{L}^*_{i} \nabla_{\Phi_\mathbf{g}} p\left(k_i | \mathbf{g}\right)  \right]\hspace{-0.5mm}-\nabla_{\Phi_\mathbf{g}} \text{KL}\left[q(\mathbf{g})\|p(\mathbf{g})\right]
$$
where $\Phi_\mathbf{g} \triangleq \text{vec}(\mathbf{C}_\mathbf{g}, \mathbf{m}_\mathbf{g})$. Then, the variational parameters $\Phi_\mathbf{g}$ can also be updated via SGA by approximating the expectation operator in $\nabla_{\Phi_\mathbf{g}} \mathcal{L}$ from sampling $\boldsymbol{\eta}_\mathbf{g} \sim q(\boldsymbol{\eta}_\mathbf{g})$.
%
%
\section{Kernel Posterior Belief and Predictive Belief of SGPR Model}
%
The optimized variational parameters $\Phi^*_\mathbf{g}$ and $\{\Phi^*_i\}_{i=1}^K$ can be computed from the above-mentioned stochastic optimization and used to induce the optimal variational distributions $q^*(\mathbf{g})$ and $q^*(\boldsymbol{\alpha}_i)$ for $i = 1, \ldots, K$ via \eqref{Lqi2}. Then, the posterior belief of the probabilistic kernel can be approximated by \emph{Monte Carlo} (MC) sampling:
\begin{equation}
\hspace{-0mm}
p(k|\mathbf{y}_\mathcal{D}) \approx q^*(k)  = 
\mathbb{E}_{q^*(\mathbf{g})}[p(k|\mathbf{g})]
 \approx  \frac{1}{S}\sum_{s=1}^S p(k|\mathbf{g}^{(s)})\hspace{-2.6mm}
\label{pk}
\end{equation}
where $\mathbf{g}^{(1)}, \ldots, \mathbf{g}^{(S)}$ are i.i.d. samples from $q^*(\mathbf{g})$. 
Such an approximated kernel posterior belief is a discrete distribution where each $q^*(k_i) \in [0,1]$ for $i = 1, \ldots, K$ can be (a) easily interpreted to identify the 
``best" kernel whose posterior probability is much larger than that of the other kernels, and (b) exploited to avoid overconfident predictions by Bayesian model averaging, as will be shown in Section~\ref{experi}.

Recall from Section~\ref{VBKS} that we approximate the posterior belief $p(\mathbf{f}_\mathcal{D}, \boldsymbol{\alpha}, k, \mathbf{g} | \mathbf{y}_\mathcal{D})$ using the variational distribution for computing the predictive belief $p(f(\mathbf{x}_*)|\mathbf{y}_\mathcal{D})$. Given
the optimal variational distributions
$q^*(\mathbf{g})$ and $q^*(\boldsymbol{\alpha}_i)$ for $i = 1, \ldots, K$, the predictive belief in~\eqref{predk} can be approximated using
\begin{equation}\label{predq}
p(f(\mathbf{x}_*)|\mathbf{y}_\mathcal{D}) \approx \sum_{i=1}^K q^*(k_i) \int p(f(\mathbf{x}_*)|\boldsymbol{\alpha}_i, k_i) \ q^*(\boldsymbol{\alpha}_i) \ \text{d}\boldsymbol{\alpha}_i
\end{equation}
which yields the following approximated predictive mean and variance for $p(f(\mathbf{x}_*)|\mathbf{y}_\mathcal{D})$:
\begin{equation}\label{mv}
\begin{array}{rl}
\mu_{\mathbf{x}_*|\mathcal{D}} \hspace{-2.4mm}&\approx \sum_{i=1}^K q^*(k_i) \ \mu_{\langle \mathbf{x}_*, i\rangle} \vspace{1mm}\\
\sigma^2_{\mathbf{x}_*|\mathcal{D}} \hspace{-2.4mm}&\approx \sum_{i=1}^K q^*(k_i)(\sigma^2_{\langle\mathbf{x}_*, i \rangle} + \mu^2_{\langle \mathbf{x}_*, i\rangle}) - \mu^2_{\mathbf{x}_*|\mathcal{D}}
\end{array}
\end{equation}
where $\mu_{\langle \mathbf{x}_*, i\rangle} $ and $\sigma^2_{\langle\mathbf{x}_*, i \rangle}$ are the predictive mean and variance of $\int p(f(\mathbf{x}_*)|\boldsymbol{\alpha}_i, k_i) \ q^*(\boldsymbol{\alpha}_i) \ \text{d}\boldsymbol{\alpha}_i$ approximated by MC sampling. The derivations of \eqref{predq}, \eqref{mv}, and the steps for computing $\mu_{\langle \mathbf{x}_*, i\rangle} $ and $\sigma^2_{\langle\mathbf{x}_*, i \rangle} $ are in\if\myproof1 Appendix~\ref{a.pred}. \fi\if\myproof0 \cite{teng2020full}. \fi

As can be seen from~\eqref{mv}, the time incurred to compute the predictive mean and variance is linear in the kernel size $K$, which is still expensive for GP prediction if $K$ is large. To overcome this issue, we can consider constructing a much smaller kernel set $\widetilde{\mathcal{K}} \subset \mathcal{K}$  including only the kernels with top-ranked posterior probabilities. Then, a new kernel posterior belief over $\widetilde{\mathcal{K}}$ can be easily computed by optimizing a new ELBO $\tilde{\mathcal{L}}(q)$ constructed from reusing the optimized local ELBOs of the kernels in $\widetilde{\mathcal{K}}$. As a result, the GP prediction can incur less time by pruning away kernels with low posterior probabilities while still maintaining the kernel uncertainty to avoid overconfident predictions.
%

In addition, one may notice that the GP predictive belief with the probabilistic kernel in \eqref{predq} is in fact performing \emph{Bayesian model averaging} (BMA) \cite{hoeting1999} over multiple GP models, each of which considers only one kernel $k_i$. Though it may seem straightforward to use BMA for GP prediction when multiple kernels are considered, our VBKS algorithm provides a principled way of doing so by deriving~\eqref{predq} from~\eqref{VBKS} using the VI framework 
and can scale the GP inference with the probabilistic kernel to big data, which is the main contribution of our work here.    
\section{Experiments and Discussion}\label{experi}
This section empirically evaluates the performance of our VBKS algorithm on two small synthetic datasets and two massive real-world datasets.
The kernel posterior belief is computed using \eqref{pk} with $S = 2000$.
The real-world experiments are performed on a Linux system with $5$ Nvidia GeForce GTX $1080$ GPUs. Stochastic optimization for VBKS is performed in a distributed manner over the $5$ GPUs using GPflow \cite{GPflow2017}. 
\subsection{Synthetic Experiments}\label{syn}
%
\begin{figure}
	\centering
	\begin{tabular}{cc}
		\hspace{-2mm}\includegraphics[scale=0.18]{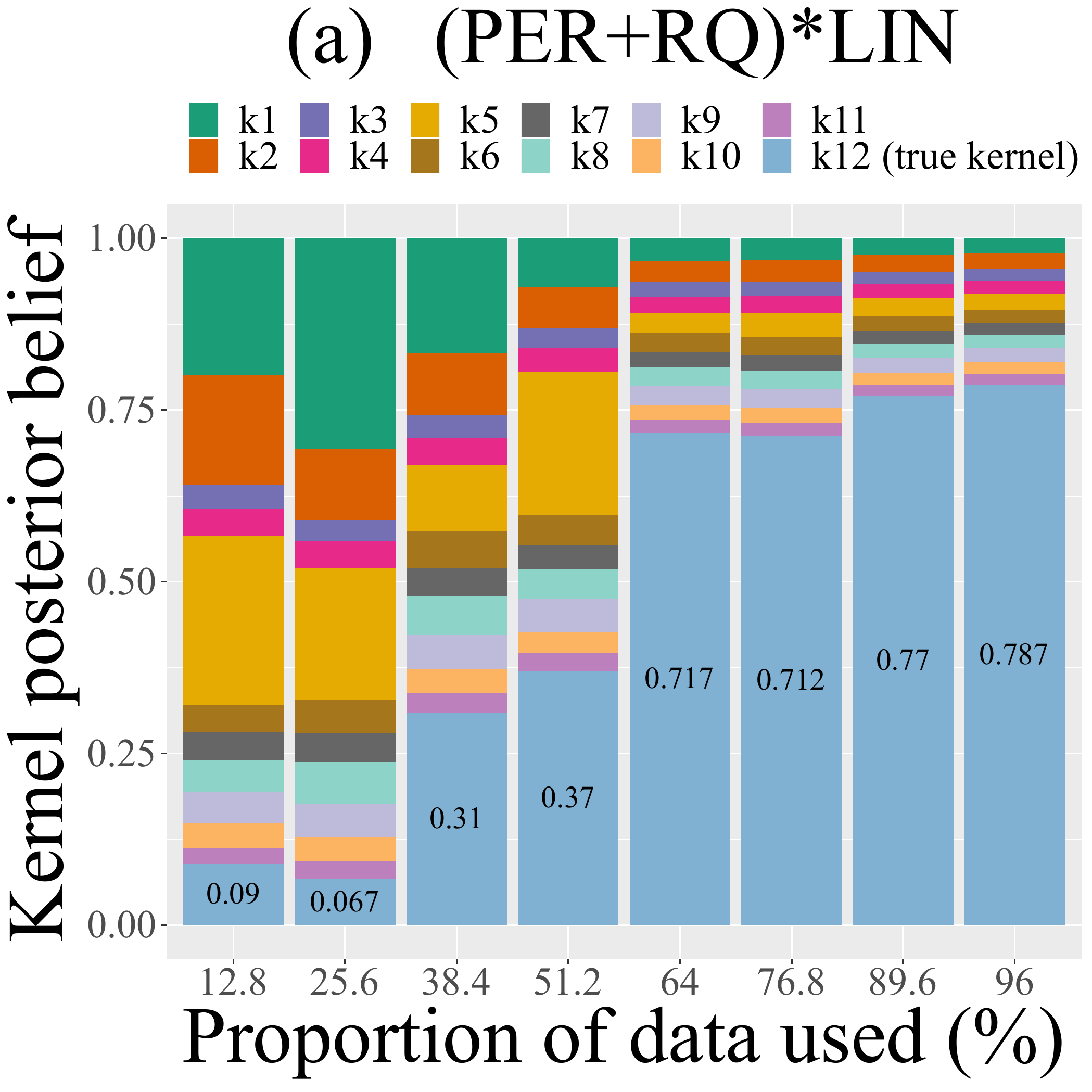} &
		\hspace{-2mm}\includegraphics[scale=0.18]{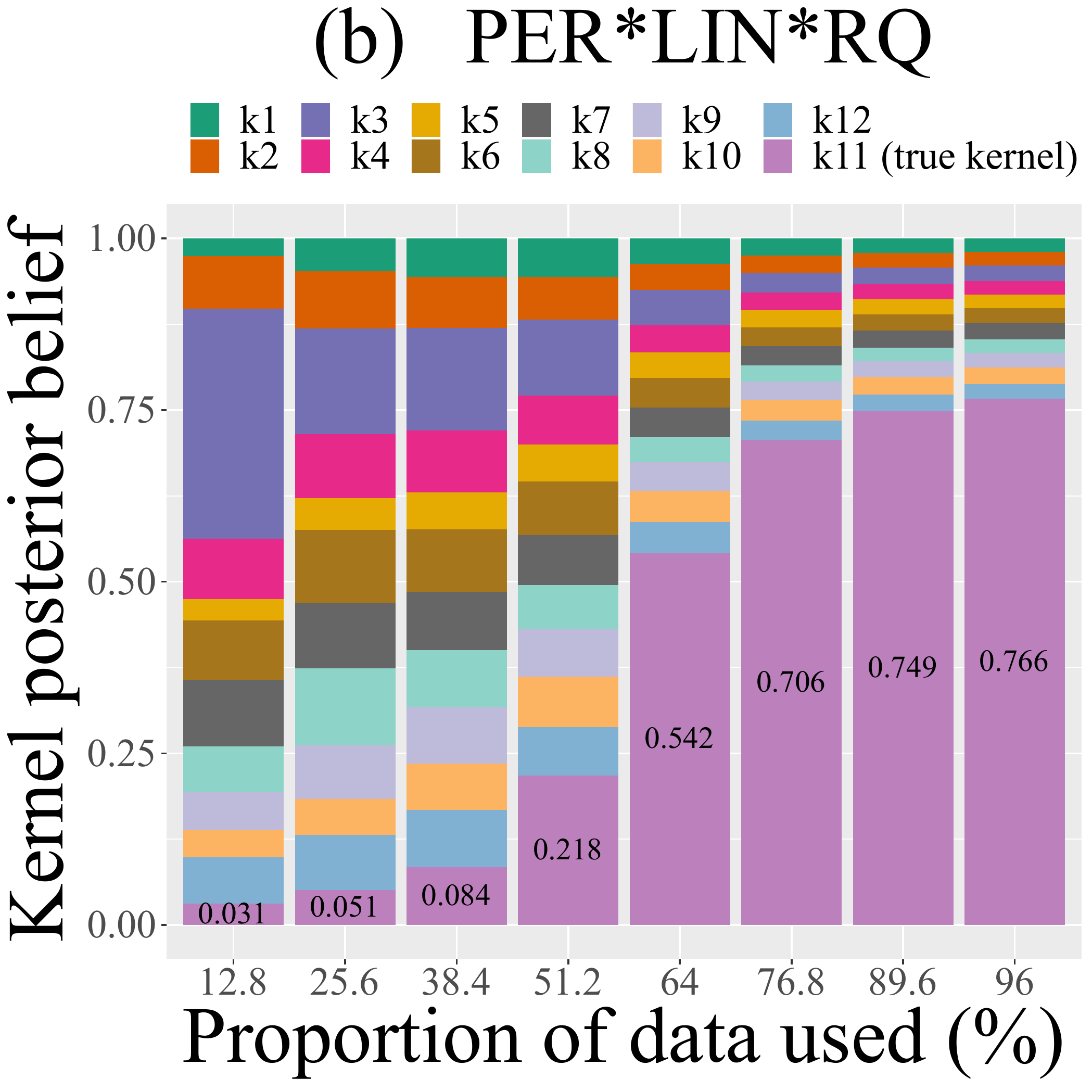} \vspace{2mm}\\
		\hspace{-2.5mm}\includegraphics[scale=0.36]{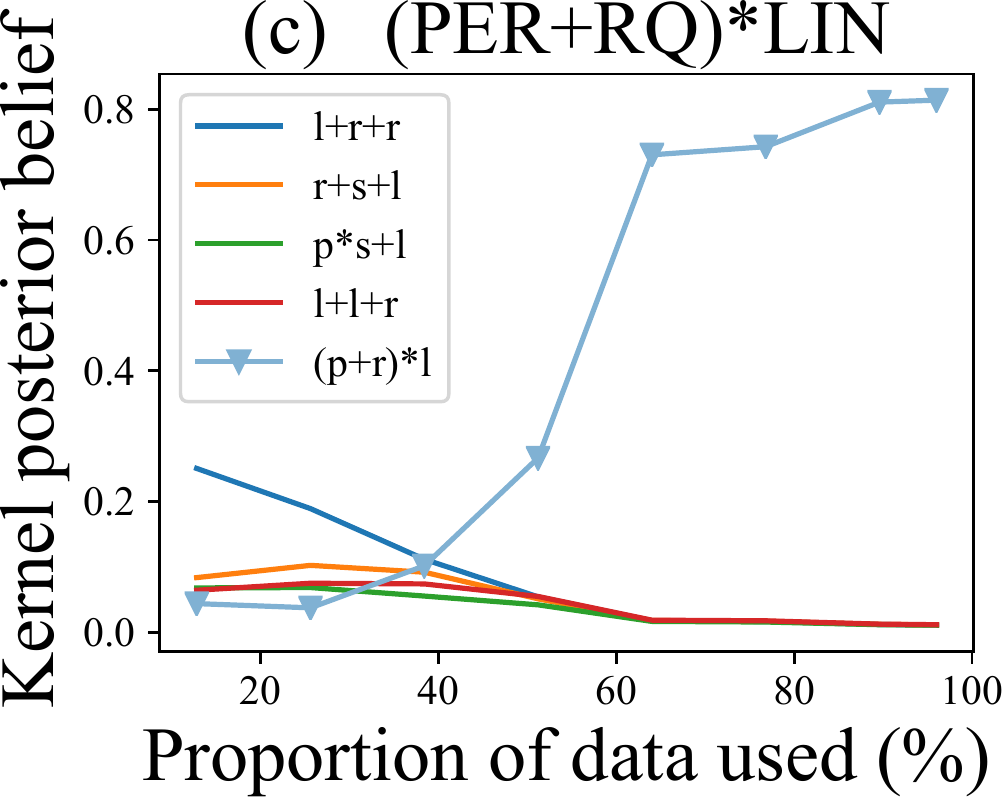} & \hspace{-3mm}\includegraphics[scale=0.36]{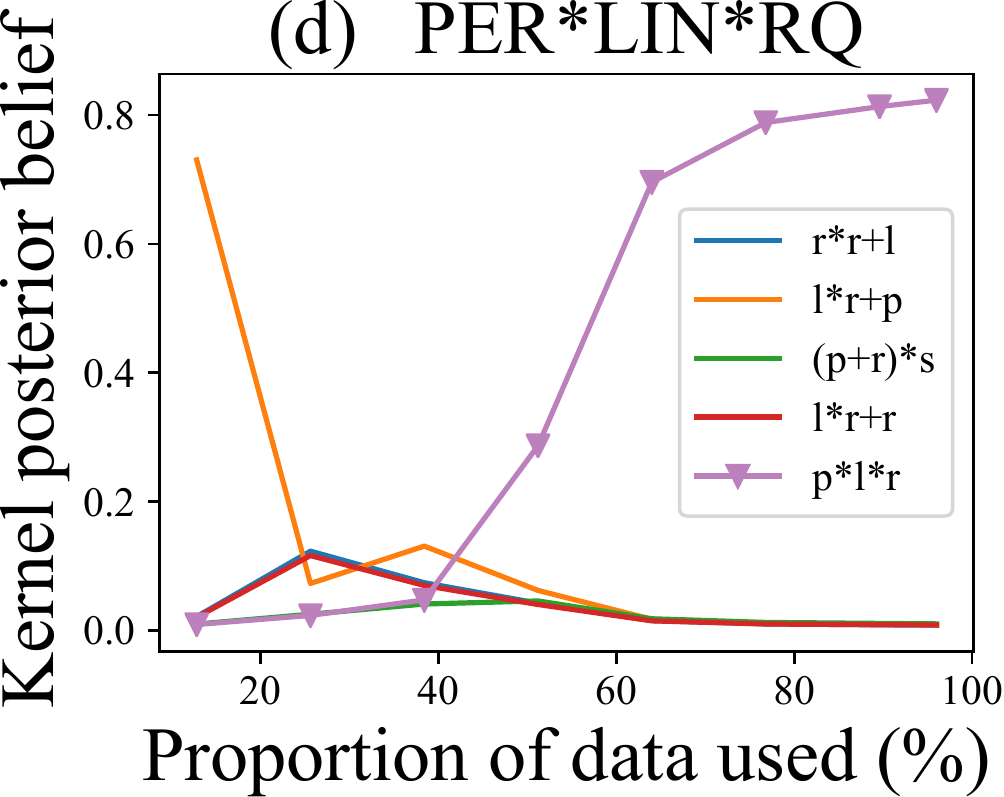}
	\end{tabular}
	\caption{Graphs of kernel posterior belief achieved by VBKS vs. proportion of data used in  stochastic optimization for two synthetic datasets with (a-b) kernel set $\mathcal{K}_{12}$ and (c-d) kernel set $\mathcal{K}_{144}$. The `s', `r', `p', and `l' in the legend denote SE, RQ, PER, and LIN kernels, respectively.}
	\label{fig.syn}
\end{figure}
We will first demonstrate the performance of our VBKS algorithm in identifying kernel(s) that can capture the underlying correlation structure of two synthetic datasets. 
%
These two synthetic datasets are generated using the respective true composite kernels $\text{PER} \times \text{LIN} \times \text{RQ}$ and $(\text{PER} + \text{RQ})\times \text{LIN} $ with fixed hyperparameters\if\myproof1 (Appendix~\ref{a.kernel}). \fi\if\myproof0 \cite{teng2020full}. \fi We set the input dimension as $d \triangleq 1$ and input domain as $\mathcal{X} \triangleq [-10, 10]$. A set $\mathcal{D}_0$ of $256$ inputs are randomly sampled from $\mathcal{X}$ and their corresponding outputs $\mathbf{y}_{\mathcal{D}_0}$ are sampled from a full-rank GP prior. Then, we sample another set $\mathcal{D}_1$ of $1000$ inputs from $\mathcal{X}$, compute their predictive mean $\boldsymbol{\mu}_{\mathcal{D}_1|\mathcal{D}_0} \triangleq (\mu_{\mathbf{x}_*|\mathcal{D}_0})_{\mathbf{x}_* \in {\mathcal{D}_1}}$
via \eqref{GPpred}, and use $(\mathcal{D}_1, \boldsymbol{\mu}_{\mathcal{D}_1|\mathcal{D}_0})$ as the synthetic dataset.
In all the synthetic experiments, we use $|\mathcal{U}| = 16$ inducing inputs
and a batch size $|\widetilde{\mathcal{D}}| = 32$ to perform the SGA update per iteration.
Two kernel sets $|\mathcal{K}_{12}| = 12$ and $|\mathcal{K}_{144}| = 144$ are used to evaluate the performance of our VBKS algorithm where $\mathcal{K}_{144}$ contains kernels constructed from the base kernels (i.e., SE, RQ, LIN, PER) by applying the grammar rules of \citeauthor{duvenaud13}~\shortcite{duvenaud13} until level $3$ while $\mathcal{K}_{12}$ includes $10$ kernels randomly sampled from $\mathcal{K}_{144}$ and the two true kernels, as shown in\if\myproof1 Appendix~\ref{a.kernel}. \fi\if\myproof0 \cite{teng2020full}. \fi The small kernel set $\mathcal{K}_{12}$ is constructed such that the posterior probabilities of all the kernels in the kernel space can be easily observed and visualized. 

Figs.~\ref{fig.syn}a and~\ref{fig.syn}b show the kernel posterior belief over $\mathcal{K}_{12}$ that is produced by our VBKS algorithm. Figs.~\ref{fig.syn}c and~\ref{fig.syn}d include only the kernels whose posterior probabilities have ever been ranked as the top two among $\mathcal{K}_{144}$ during stochastic optimization for VBKS.
%
It can be observed in all the experiments that the posterior probability of the true kernel is small at the beginning and, with a growing data size, is increased by our VBKS algorithm to be around $0.8$ which is much larger than that of the other kernels.
Though the SGPR model used by our VBKS algorithm produces a low-rank approximation of the true covariance structure of the data using a small set of inducing variables, our algorithm can find the kernel that correctly interprets the underlying correlation structure.
It can be observed from Figs.~\ref{fig.syn}a and~\ref{fig.syn}b that the kernel uncertainty is large (i.e., no kernel has a much larger posterior probability than the others) when less than half of the data is used in these experiments. In such cases, the high kernel uncertainty shows that the current observed data is insufficient in identifying a single kernel that fits the true correlation structure much better than the other kernels. This implies the need to collect more data (as has been done in the experiments) or perform GP prediction with the kernel uncertainty. 
%
%
\subsection{Real-World Experiments}
%
This section empirically evaluates the performance and time efficiency of our VBKS algorithm on two massive real-world datasets: (a) Swissgrid dataset\footnote{\url{https://www.swissgrid.ch}} contains $210,336$ records of the total energy consumed by end users in the Swiss control block from January $1$, $2009$ to December $31$, $2015$ in every $15$ minutes, and (b) \textit{indoor environmental quality} (IEQ) dataset\footnote{\url{http://db.csail.mit.edu/labdata/labdata.html}}
contains temperature ($^\circ$C) taken in every $31$ seconds between February $28$ and April $5$, $2004$ by $54$ sensors deployed in the Intel Berkeley Research lab.

The candidate kernel set $\mathcal{K}_{144}$ is used in the experiments for both datasets. The predictive performance of our VBKS algorithm is obtained using BMA over $10$ kernels with top-ranked posterior probabilities and compared with that of (a) \emph{Bayesian optimization} (BO) for kernel selection \cite{malkomes16b} which automatically finds the kernel with the highest model evidence using BO, (b) random algorithm which randomly selects a kernel from $\mathcal{K}_{144}$ at the beginning of each experiment, and (c) VBKS performing GP prediction with a single kernel that yields the highest posterior probability (VBKS-s).
For each dataset, $20,000$ observations are randomly selected to form the test set $\mathcal{T}$. The \emph{root mean squared error} (RMSE) metric $\sqrt{|\mathcal{T}|^{-1}\sum_{\mathbf{x}_* \in \mathcal{T}}(y(\mathbf{x}_*) - \mu_{\mathbf{x}_*|\mathcal{D}})^2}$ is used to evaluate the predictive performance of the tested algorithms. 
The RMSEs of the VBKS(-s) and random algorithms are averaged over $10$ and $50$ independent runs, respectively.
The error bars are computed in the form of standard deviation.
To save some time from sampling $q(\boldsymbol{\theta}_i)$, the hyperparameters used by all tested algorithms are point estimates by setting $\mathbf{C}_{\boldsymbol{\theta}_i}$ as a zero matrix.
\begin{figure}
	\centering
	\begin{tabular}{cc}
		\hspace{-2mm} \includegraphics[scale=0.37]{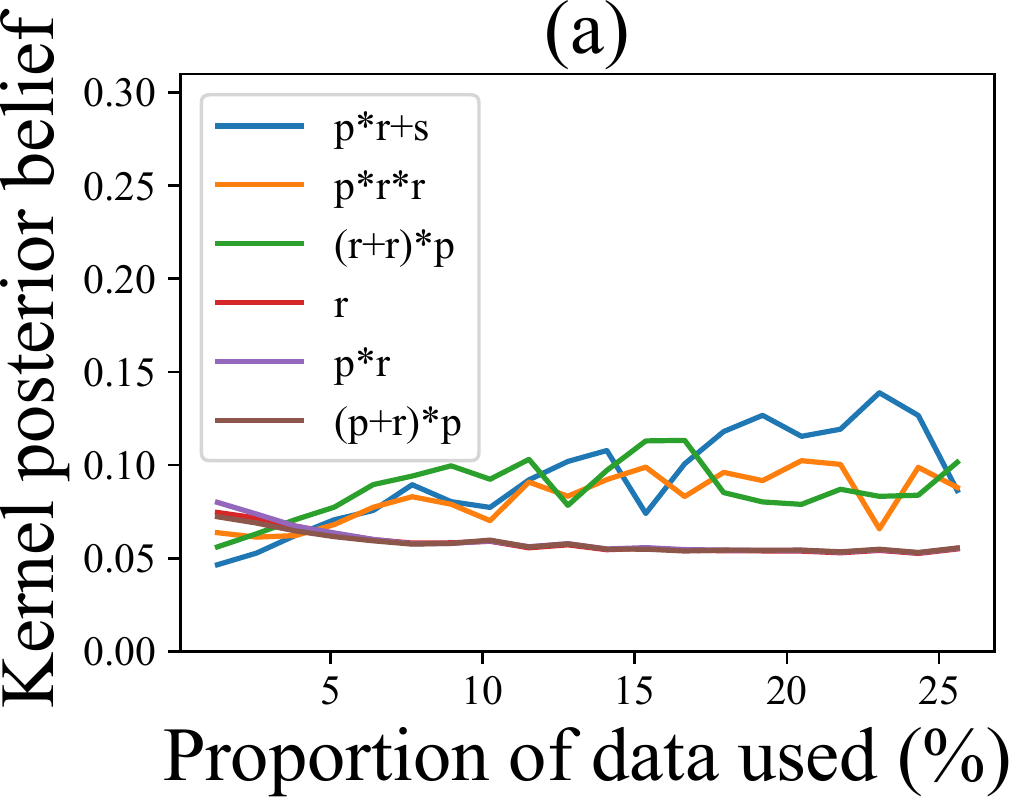} & \hspace{-3mm}\includegraphics[scale=0.38]{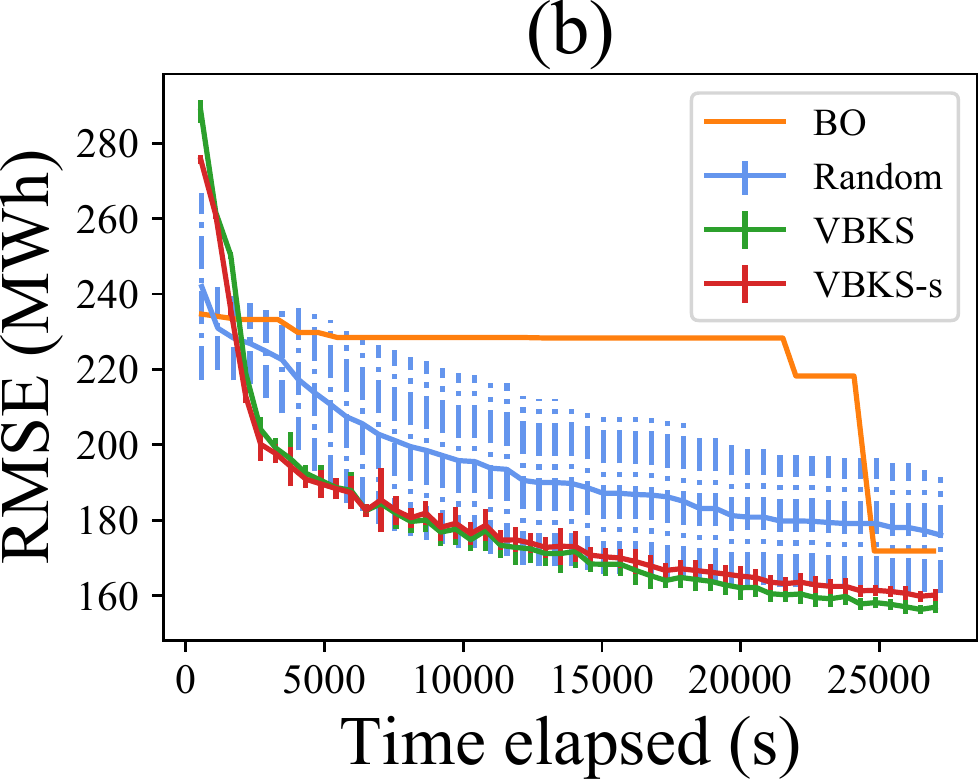}
	\end{tabular}
	\caption{Graphs of (a) kernel posterior belief over selected kernels that is produced by our VBKS algorithm in one run vs. proportion of data used in stochastic optimization, and (b) RMSE vs. time incurred by  tested algorithms for Swissgrid dataset.}
	\label{fig.swiss}
\end{figure}
\begin{figure}
	\centering
	\begin{tabular}{cc}
		\hspace{-2mm} \includegraphics[scale=0.37]{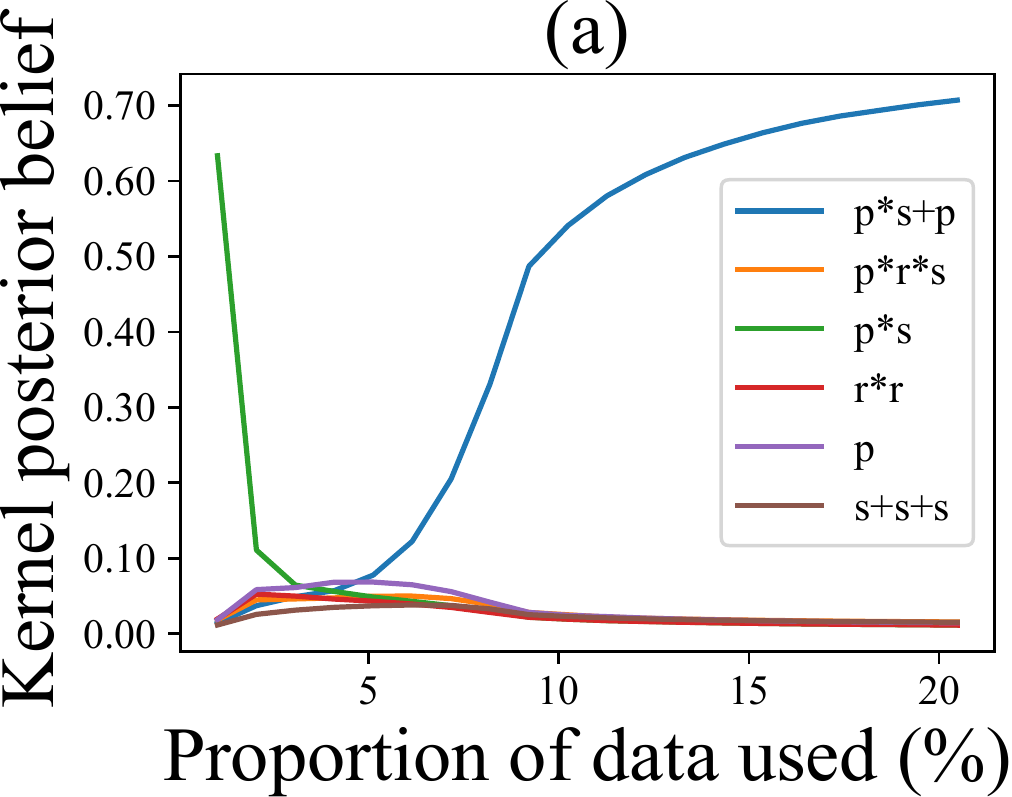} & \hspace{-3mm}\includegraphics[scale=0.38]{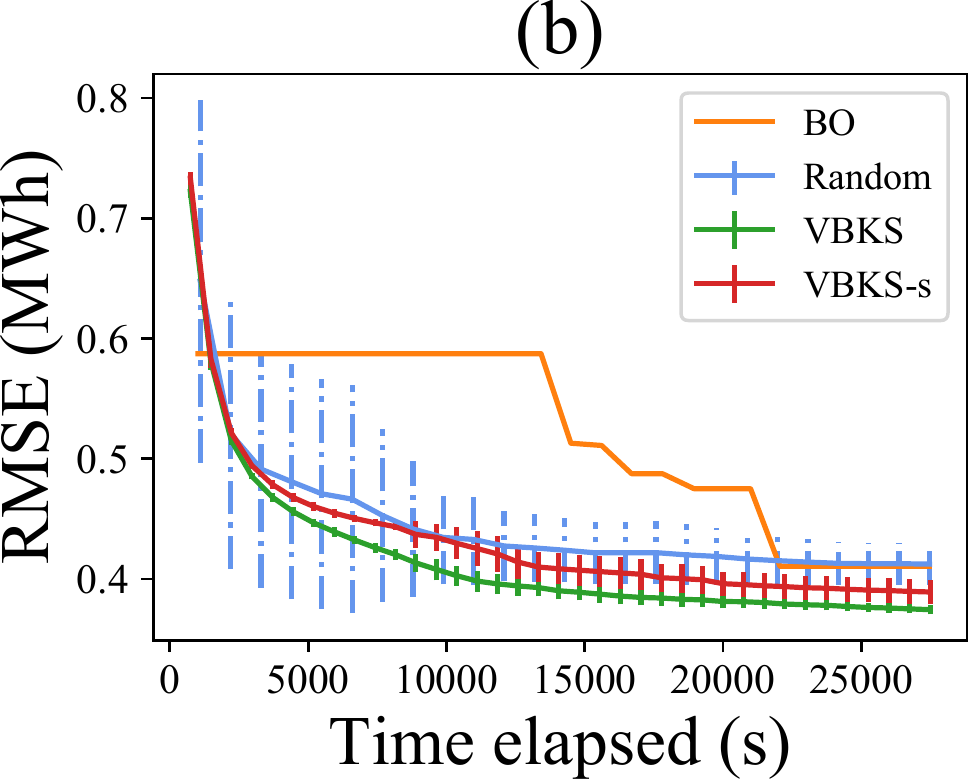}
	\end{tabular}
	\caption{Graphs of (a) kernel posterior belief over selected kernels that is produced by our VBKS algorithm in one run vs. proportion of data used in stochastic optimization, and (b) RMSE vs. time incurred by  tested algorithms for IEQ dataset.}
	\label{fig.IEQ}
\end{figure}
\subsubsection{Swissgrid Dataset.}
We use $|\mathcal{U}| = 800$ inducing inputs and a batch size $|\widetilde{\mathcal{D}}| = 128$ for SGA update per iteration. 
The kernel posterior belief produced by our VBKS algorithm is evaluated after every 1.28\% of data are used, as shown in Fig.~\ref{fig.swiss}a.
For clarity, we only visualize the kernels whose posterior probabilities have ever been ranked as the top three during stochastic optimization for VBKS. 
It can be observed that a group of three kernels (rather than a single kernel) achieve comparable posterior probabilities which are larger than that of the other kernels.
%
This implies that the uncertainty among the kernels need to be considered when the selected kernels are used to interpret the correlation structure or perform GP prediction, which is a key benefit of using our VBKS algorithm.
The predictive performance of the tested algorithms for Swissgrid dataset is shown in Fig.~\ref{fig.swiss}b: Both the VBKS and VBKS-s algorithms converge faster to a smaller RMSE than the other tested algorithms. VBKS performs better than VBKS-s since BMA can benefit from different kernels in modeling the data when no kernel truly stands out.
The BO algorithm performs poorly because it has to approximate the distance between kernels using a small subset of data (i.e., of size $200$ here).\footnote{We also tried a larger subset of data of size $500$ to better approximate the kernel distances in BO, which yields similar predictive performance but incurs much more time.} 
If the small subset of data is not large enough to approximate the kernel distances well, the BO performance will degrade, which is the case in our experiments.
%
\subsubsection{IEQ Dataset.}
In this experiment, the time and locations of the sensors for producing the temperature readings are used jointly as the input (i.e., $d = 3$). The first $1$ million valid data points are selected for our experiments. We use $|\mathcal{U}|=1000$ inducing inputs and a batch size $|\widetilde{\mathcal{D}}| = 512$ for SGA update per iteration. Fig.~\ref{fig.IEQ} shows the kernel posterior belief and RMSE of the tested algorithms for IEQ dataset. Different from the results in Fig.~\ref{fig.swiss}a for Swissgrid dataset, Fig.~\ref{fig.IEQ}a shows that the posterior probability of a specific kernel (i.e., $\text{PER}\times \text{SE}+\text{PER}$) increases fast to be much larger than that of the other kernels.
However, it can be observed from Fig.~\ref{fig.IEQ}b that VBKS still outperforms VBKS-s because the training of GP model using $\text{PER}\times \text{SE}+\text{PER}$ overfits to the training data and our VBKS algorithm with BMA helps to reduce the overfitting.
VBKS also converges to a smaller RMSE than all other tested algorithms, as shown in Fig.~\ref{fig.IEQ}b.
%
In addition, for both Swissgrid and IEQ datasets, we observe that VBKS has consistently achieved smaller RMSE than VBKS-s (albeit slightly) in all $10$ independent runs, which demonstrates the benefit of applying BMA. The results of all these $10$ runs (instead of the averaged RMSE) are in\if\myproof1 Appendix \ref{a.result}.  \fi\if\myproof0 \cite{teng2020full}. \fi 
\subsubsection{Scalability.}
\begin{figure}
	\centering
	\includegraphics[scale=0.45]{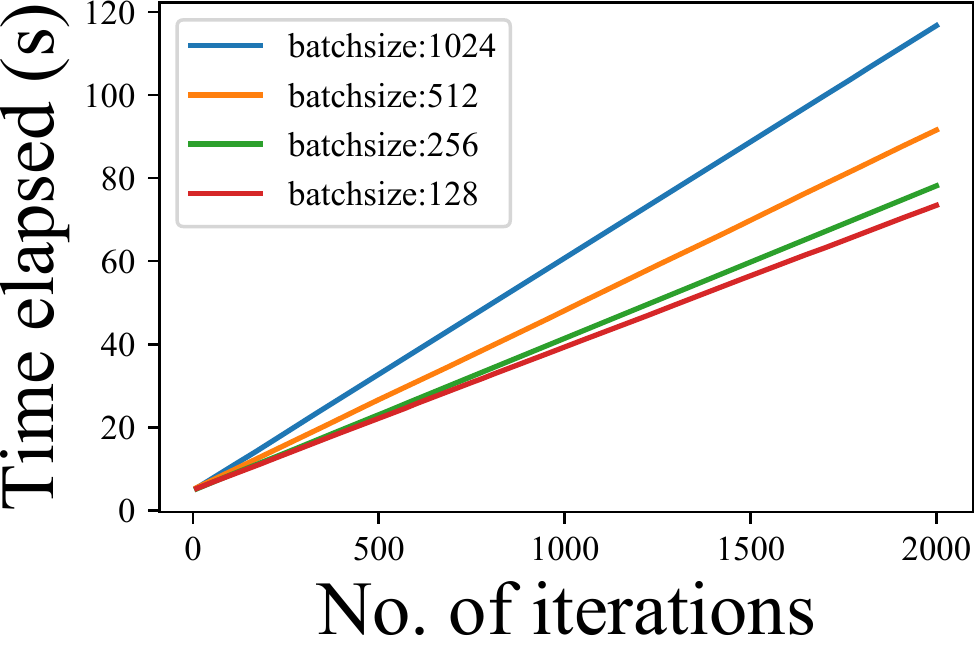} 
	\caption{Graph of total incurred time of our VBKS algorithm vs. number of iterations of SGA updates with $|\mathcal{U}| = 800$ and varying batch sizes $|\widetilde{\mathcal{D}}|$.} 
	\label{fig.time}
\end{figure}

Fig.~\ref{fig.time} shows the time efficiency of our VBKS algorithm for different batch sizes for SGA update per iteration. As expected, the total time incurred by VBKS increases linearly in the number of iterations of SGA updates.
%
%
\section{Conclusion}
This paper describes a novel VBKS algorithm for SGPR models that considers a probabilistic kernel and exploits the kernel uncertainty to avoid overconfident predictions. A stochastic optimization method for VBKS is proposed for learning the kernel variational distribution together with the other variational variables (i.e., inducing variables and kernel hyperparameters). Our VBKS algorithm achieves scalability in the kernel size $K$ by decomposing the variational lower bound into an additive form such that each additive term (i.e., local ELBO) depends on the variational variables of only one kernel and can thus be maximized independently. Then, each additive local ELBO is optimized via SGA by sampling from both the variational distribution and the data, which scales to big data since it incurs constant time per iteration of SGA update. The predictive performance of our VBKS algorithm with BMA is shown to outperform the other tested kernel selection algorithms.
A limitation of VBKS is that it uses a finite and fixed kernel space, which does not allow flexible exploration. In our future work, we will consider expanding the kernel space during stochastic optimization according to the intermediate learning outcomes and extending VBKS to operate with an infinite kernel space and a deep GP model~\cite{haibin19}. 
%
%
\subsubsection{Acknowledgments.} 
This research is supported by the National Research Foundation, Prime Minister's Office, Singapore under its Campus for Research Excellence and Technological Enterprise (CREATE) program, Singapore-MIT Alliance for Research and Technology (SMART) Future Urban Mobility (FM) IRG, the Singapore Ministry of Education Academic Research Fund Tier $2$, MOE$2016$-T$2$-$2$-$156$, and the Joint Funds of the National Natural Science Foundation of China under Key Program Grant U$1713212$.
 
\bibliographystyle{aaai}
\bibliography{BKS}

\begin{thebibliography}{}

\bibitem[\protect\citeauthoryear{Benton \bgroup et al\mbox.\egroup
  }{2019}]{benton2019}
Benton, G.~W.; Maddox, W.~J.; Salkey, J.~P.; Albinati, J.; and Wilson, A.~G.
\newblock 2019.
\newblock Function-space distributions over kernels.
\newblock In {\em Proc. {NeurIPS}}.

\bibitem[\protect\citeauthoryear{Chen \bgroup et al\mbox.\egroup
  }{2012}]{LowUAI12}
Chen, J.; Low, K.~H.; Tan, C. K.-Y.; Oran, A.; Jaillet, P.; Dolan, J.~M.; and
  Sukhatme, G.~S.
\newblock 2012.
\newblock Decentralized data fusion and active sensing with mobile sensors for
  modeling and predicting spatiotemporal traffic phenomena.
\newblock In {\em Proc. {UAI}},  163--173.

\bibitem[\protect\citeauthoryear{Chen \bgroup et al\mbox.\egroup
  }{2013}]{LowUAI13}
Chen, J.; Cao, N.; Low, K.~H.; Ouyang, R.; Tan, C. K.-Y.; and Jaillet, P.
\newblock 2013.
\newblock Parallel {Gaussian} process regression with low-rank covariance
  matrix approximations.
\newblock In {\em Proc. {UAI}},  152--161.

\bibitem[\protect\citeauthoryear{Chen \bgroup et al\mbox.\egroup
  }{2015}]{LowTASE15}
Chen, J.; Low, K.~H.; Jaillet, P.; and Yao, Y.
\newblock 2015.
\newblock Gaussian process decentralized data fusion and active sensing for
  spatiotemporal traffic modeling and prediction in mobility-on-demand systems.
\newblock {\em {IEEE} Transactions on Automation Science and Engineering}
  12(3):901--921.

\bibitem[\protect\citeauthoryear{Duvenaud \bgroup et al\mbox.\egroup
  }{2013}]{duvenaud13}
Duvenaud, D.; Lloyd, J.~R.; Grosse, R.; Tenenbaum, J.~B.; and Ghahramani, Z.
\newblock 2013.
\newblock Structure discovery in nonparametric regression through compositional
  kernel search.
\newblock In {\em Proc. {ICML}},  1166--1174.

\bibitem[\protect\citeauthoryear{Gal and Turner}{2015}]{gal2015}
Gal, Y., and Turner, R.
\newblock 2015.
\newblock Improving the {G}aussian process sparse spectrum approximation by
  representing uncertainty in frequency inputs.
\newblock In {\em Proc. {ICML}},  655--664.

\bibitem[\protect\citeauthoryear{Hensman \bgroup et al\mbox.\egroup
  }{2015}]{hensman2015}
Hensman, J.; Matthews, A.~G.; Filippone, M.; and Ghahramani, Z.
\newblock 2015.
\newblock {MCMC} for variationally sparse {G}aussian processes.
\newblock In {\em Proc. {NeurIPS}},  1648--1656.

\bibitem[\protect\citeauthoryear{Hoang, Hoang, and Low}{2015}]{Hoang2015}
Hoang, T.~N.; Hoang, Q.~M.; and Low, K.~H.
\newblock 2015.
\newblock A unifying framework of anytime sparse {G}aussian process regression
  models with stochastic variational inference for big data.
\newblock In {\em Proc. {ICML}},  569--578.

\bibitem[\protect\citeauthoryear{Hoang, Hoang, and Low}{2016}]{HoangICML16}
Hoang, T.~N.; Hoang, Q.~M.; and Low, K.~H.
\newblock 2016.
\newblock A distributed variational inference framework for unifying parallel
  sparse {Gaussian} process regression models.
\newblock In {\em Proc. {ICML}},  382--391.

\bibitem[\protect\citeauthoryear{Hoang, Hoang, and Low}{2017}]{hoang2017}
Hoang, Q.~M.; Hoang, T.~N.; and Low, K.~H.
\newblock 2017.
\newblock A generalized stochastic variational {B}ayesian hyperparameter
  learning framework for sparse spectrum {G}aussian process regression.
\newblock In {\em Proc. {AAAI}},  2007--2014.

\bibitem[\protect\citeauthoryear{Hoeting \bgroup et al\mbox.\egroup
  }{1999}]{hoeting1999}
Hoeting, J.~A.; Madigan, D.; Raftery, A.~E.; and Volinsky, C.~T.
\newblock 1999.
\newblock Bayesian model averaging: a tutorial.
\newblock {\em Statistical science}  382--401.

\bibitem[\protect\citeauthoryear{Kim and Teh}{2018}]{kim2018}
Kim, H., and Teh, Y.~W.
\newblock 2018.
\newblock Scaling up the automatic statistician: Scalable structure discovery
  using {G}aussian processes.
\newblock In {\em Proc. {AISTATS}}.

\bibitem[\protect\citeauthoryear{L{\'a}zaro-Gredilla \bgroup et al\mbox.\egroup
  }{2010}]{lazaro2010}
L{\'a}zaro-Gredilla, M.; Qui{\~n}onero-Candela, J.; Rasmussen, C.~E.; and
  Figueiras-Vidal, A.~R.
\newblock 2010.
\newblock Sparse spectrum {G}aussian process regression.
\newblock {\em JMLR} 11:1865--1881.

\bibitem[\protect\citeauthoryear{Lloyd \bgroup et al\mbox.\egroup
  }{2014}]{lloyd2014}
Lloyd, J.~R.; Duvenaud, D.; Grosse, R.; Tenenbaum, J.~B.; and Ghahramani, Z.
\newblock 2014.
\newblock Automatic construction and natural-language description of
  nonparametric regression models.
\newblock In {\em Proc. {AAAI}}.

\bibitem[\protect\citeauthoryear{Low \bgroup et al\mbox.\egroup
  }{2015a}]{LowDyDESS15}
Low, K.~H.; Chen, J.; Hoang, T.~N.; Xu, N.; and Jaillet, P.
\newblock 2015a.
\newblock Recent advances in scaling up {Gaussian} process predictive models
  for large spatiotemporal data.
\newblock In Ravela, S., and Sandu, A., eds., {\em Proc. Dynamic Data-driven
  Environmental Systems Science Conference ({DyDESS'14})}. LNCS 8964, Springer.

\bibitem[\protect\citeauthoryear{Low \bgroup et al\mbox.\egroup
  }{2015b}]{LowAAAI15}
Low, K.~H.; Yu, J.; Chen, J.; and Jaillet, P.
\newblock 2015b.
\newblock Parallel {Gaussian} process regression for big data: Low-rank
  representation meets {M}arkov approximation.
\newblock In {\em Proc. {AAAI}}.

\bibitem[\protect\citeauthoryear{Lu \bgroup et al\mbox.\egroup }{2018}]{lu2018}
Lu, X.; Gonzalez, J.; Dai, Z.; and Lawrence, N.
\newblock 2018.
\newblock Structured variationally auto-encoded optimization.
\newblock In {\em Proc. {ICML}},  3273--3281.

\bibitem[\protect\citeauthoryear{Malkomes and Garnett}{2015}]{malkomes16b}
Malkomes, G., and Garnett, R.
\newblock 2015.
\newblock Active structure discovery for {G}aussian processes.
\newblock In {\em {ICML} Workshop on {AutoML}}.

\bibitem[\protect\citeauthoryear{Malkomes, Schaff, and
  Garnett}{2016}]{malkomes2016}
Malkomes, G.; Schaff, C.; and Garnett, R.
\newblock 2016.
\newblock Bayesian optimization for automated model selection.
\newblock In {\em Proc. {NeurIPS}},  2900--2908.

\bibitem[\protect\citeauthoryear{Matthews \bgroup et al\mbox.\egroup
  }{2017}]{GPflow2017}
Matthews, A. G. d.~G.; {van der Wilk}, M.; Nickson, T.; Fujii, K.;
  {Boukouvalas}, A.; {Le{\'o}n-Villagr{\'a}}, P.; Ghahramani, Z.; and Hensman,
  J.
\newblock 2017.
\newblock {{GP}flow: A {G}aussian process library using {T}ensor{F}low}.
\newblock {\em JMLR} 18(40):1--6.

\bibitem[\protect\citeauthoryear{Oliva \bgroup et al\mbox.\egroup
  }{2016}]{oliva2016}
Oliva, J.~B.; Dubey, A.; Wilson, A.~G.; P{\'o}czos, B.; Schneider, J.; and
  Xing, E.~P.
\newblock 2016.
\newblock Bayesian nonparametric kernel-learning.
\newblock In {\em Proc. {AISTATS}},  1078--1086.

\bibitem[\protect\citeauthoryear{Qui{\~n}onero-Candela and
  Rasmussen}{2005}]{Quinonero2005}
Qui{\~n}onero-Candela, J., and Rasmussen, C.~E.
\newblock 2005.
\newblock A unifying view of sparse approximate {G}aussian process regression.
\newblock {\em JMLR} 6:1939--1959.

\bibitem[\protect\citeauthoryear{Rasmussen and Williams}{2006}]{Rasmussen2006}
Rasmussen, C.~E., and Williams, C.~K.
\newblock 2006.
\newblock {\em Gaussian processes for machine learning}.
\newblock MIT Press.

\bibitem[\protect\citeauthoryear{Titsias and
  L{\'a}zaro-Gredilla}{2013}]{Titsias2013}
Titsias, M.~K., and L{\'a}zaro-Gredilla, M.
\newblock 2013.
\newblock Variational inference for {M}ahalanobis distance metrics in
  {G}aussian process regression.
\newblock In {\em Proc. {NeurIPS}},  279--287.

\bibitem[\protect\citeauthoryear{Titsias and
  L{\'a}zaro-Gredilla}{2014}]{titsias2014}
Titsias, M.~K., and L{\'a}zaro-Gredilla, M.
\newblock 2014.
\newblock Doubly stochastic variational {B}ayes for non-conjugate inference.
\newblock In {\em Proc. {ICML}},  1971--1979.

\bibitem[\protect\citeauthoryear{Titsias}{2009}]{Titsias2009}
Titsias, M.~K.
\newblock 2009.
\newblock Variational learning of inducing variables in sparse {G}aussian
  processes.
\newblock In {\em Proc. {AISTATS}}.

\bibitem[\protect\citeauthoryear{Wilson and Adams}{2013}]{wilson2013}
Wilson, A.~G., and Adams, R.~P.
\newblock 2013.
\newblock Gaussian process kernels for pattern discovery and extrapolation.
\newblock In {\em Proc. {ICML}}.

\bibitem[\protect\citeauthoryear{Xu \bgroup et al\mbox.\egroup
  }{2014}]{LowAAAI14}
Xu, N.; Low, K.~H.; Chen, J.; Lim, K.~K.; and {\"{O}zg\"{u}l}, E.~B.
\newblock 2014.
\newblock {GP-Localize}: Persistent mobile robot localization using online
  sparse {Gaussian} process observation model.
\newblock In {\em Proc. {AAAI}},  2585--2592.

\bibitem[\protect\citeauthoryear{Yu \bgroup et al\mbox.\egroup
  }{2019a}]{haibin19}
Yu, H.; Chen, Y.; Dai, Z.; Low, K.~H.; and Jaillet, P.
\newblock 2019a.
\newblock Implicit posterior variational inference for deep {Gaussian}
  processes.
\newblock In {\em Proc. NeurIPS}.

\bibitem[\protect\citeauthoryear{Yu \bgroup et al\mbox.\egroup
  }{2019b}]{Yu2019}
Yu, H.; Hoang, T.~N.; Low, K.~H.; and Jaillet, P.
\newblock 2019b.
\newblock Stochastic variational inference for {B}ayesian sparse {G}aussian
  process regression.
\newblock In {\em Proc. {IJCNN}}.

\end{thebibliography}

\if \myproof1
\clearpage

\appendix
\section{Derivation of \eqref{elbo}}\label{a.elbo}
Since $p(\mathbf{y}_\mathcal{D}) = p(\mathbf{y}_\mathcal{D}, \mathbf{f}_\mathcal{D}, \boldsymbol{\alpha}, k, \mathbf{g}) / p(\mathbf{f}_\mathcal{D}, \boldsymbol{\alpha}, k, \mathbf{g} | \mathbf{y}_\mathcal{D})$ for any $\mathbf{f}_\mathcal{D}, \boldsymbol{\alpha}, k, \text{and}\ \mathbf{g}$, 
\begin{equation}\label{y}
\log p(\mathbf{y}_\mathcal{D}) = \log \frac{p(\mathbf{y}_\mathcal{D}, \mathbf{f}_\mathcal{D}, \boldsymbol{\alpha}, k, \mathbf{g})}{ p(\mathbf{f}_\mathcal{D}, \boldsymbol{\alpha}, k, \mathbf{g} | \mathbf{y}_\mathcal{D})}\ .
\end{equation}
Let $q(\mathbf{f}_\mathcal{D}, \boldsymbol{\alpha}, k, \mathbf{g})$ be an arbitrary probability density function. Then, we can take an expectation w.r.t.~$q(\mathbf{f}_\mathcal{D}, \boldsymbol{\alpha}, k, \mathbf{g})$ on both side of \eqref{y}, which yields
$$
\hspace{-1.7mm}
\begin{array}{l}
\log p(\mathbf{y}_\mathcal{D})  \vspace{1mm}\\
=\displaystyle \mathbb{E}_{q(\mathbf{f}_\mathcal{D}, \boldsymbol{\alpha}, k, \mathbf{g})} \left[\log \frac{p(\mathbf{y}_\mathcal{D}, \mathbf{f}_\mathcal{D}, \boldsymbol{\alpha}, k, \mathbf{g})}{ p(\mathbf{f}_\mathcal{D}, \boldsymbol{\alpha}, k, \mathbf{g} | \mathbf{y}_\mathcal{D})} \right]
 \vspace{1mm}\\
=\displaystyle \mathbb{E}_{q(\mathbf{f}_\mathcal{D}, \boldsymbol{\alpha}, k, \mathbf{g})}
\left[
\log \left( \frac{p(\mathbf{y}_\mathcal{D}, \mathbf{f}_\mathcal{D}, \boldsymbol{\alpha}, k, \mathbf{g})}
{ q(\mathbf{f}_\mathcal{D}, \boldsymbol{\alpha}, k, \mathbf{g})} \frac{q(\mathbf{f}_\mathcal{D}, \boldsymbol{\alpha}, k, \mathbf{g})}
{ p(\mathbf{f}_\mathcal{D}, \boldsymbol{\alpha}, k, \mathbf{g} | \mathbf{y}_\mathcal{D})}\right)
\right] \vspace{1mm}\\
=\displaystyle \mathbb{E}_{q(\mathbf{f}_\mathcal{D}, \boldsymbol{\alpha}, k, \mathbf{g})}
\left[
\log \frac{p(\mathbf{y}_\mathcal{D} |\mathbf{f}_\mathcal{D}, \boldsymbol{\alpha}, k, \mathbf{g})p(\mathbf{f}_\mathcal{D}, \boldsymbol{\alpha}, k, \mathbf{g})}
{ q(\mathbf{f}_\mathcal{D}, \boldsymbol{\alpha}, k, \mathbf{g})}
\right] \vspace{1mm}\\
\quad \displaystyle +\ \mathbb{E}_{q(\mathbf{f}_\mathcal{D}, \boldsymbol{\alpha}, k, \mathbf{g})} 
 \left[\log\frac{q(\mathbf{f}_\mathcal{D}, \boldsymbol{\alpha}, k, \mathbf{g})}
{ p(\mathbf{f}_\mathcal{D}, \boldsymbol{\alpha}, k, \mathbf{g} | \mathbf{y}_\mathcal{D})}\right]\vspace{1mm}\\
=\displaystyle \mathbb{E}_{q(\mathbf{f}_\mathcal{D}, \boldsymbol{\alpha}, k, \mathbf{g})}
\left[\log p(\mathbf{y}_\mathcal{D} | \mathbf{f}_\mathcal{D})\right] \vspace{1mm}\\
\quad \displaystyle -\ \mathbb{E}_{q(\mathbf{f}_\mathcal{D}, \boldsymbol{\alpha}, k, \mathbf{g})}
\left[ \log\frac{q(\mathbf{f}_\mathcal{D}, \boldsymbol{\alpha}, k, \mathbf{g})}
{ p(\mathbf{f}_\mathcal{D}, \boldsymbol{\alpha}, k, \mathbf{g})}
\right] \vspace{1mm}\\
\quad  \displaystyle +\ \mathbb{E}_{q(\mathbf{f}_\mathcal{D}, \boldsymbol{\alpha}, k, \mathbf{g})} 
\left[\log\frac{q(\mathbf{f}_\mathcal{D}, \boldsymbol{\alpha}, k, \mathbf{g})}
{ p(\mathbf{f}_\mathcal{D}, \boldsymbol{\alpha}, k, \mathbf{g} | \mathbf{y}_\mathcal{D})}\right] \vspace{1mm}\\
=\displaystyle \mathbb{E}_{q(\mathbf{f}_\mathcal{D}, \boldsymbol{\alpha}, k, \mathbf{g})}
\left[\log p(\mathbf{y}_\mathcal{D} | \mathbf{f}_\mathcal{D})\right] \vspace{1mm}\\
\quad -\ \text{KL}\left[q(\mathbf{f}_\mathcal{D}, \boldsymbol{\alpha}, k, \mathbf{g}) \|
p(\mathbf{f}_\mathcal{D}, \boldsymbol{\alpha}, k, \mathbf{g})
\right] \vspace{1mm}\\
\quad\displaystyle +\ \text{KL}
\left[q(\mathbf{f}_\mathcal{D}, \boldsymbol{\alpha}, k, \mathbf{g}) \|
p(\mathbf{f}_\mathcal{D}, \boldsymbol{\alpha}, k, \mathbf{g} | \mathbf{y}_\mathcal{D})\right] \vspace{1mm}\\
=\displaystyle \mathcal{L}(q) + \displaystyle \text{KL}
\left[q(\mathbf{f}_\mathcal{D}, \boldsymbol{\alpha}, k, \mathbf{g}) \|
p(\mathbf{f}_\mathcal{D}, \boldsymbol{\alpha}, k, \mathbf{g} | \mathbf{y}_\mathcal{D})\right]
\end{array}
$$
where the fourth equality is due to the fact that $\mathbf{y}_\mathcal{D}$ is conditionally independent of $(\boldsymbol{\alpha}, k, \mathbf{g})$ given $\mathbf{f}_\mathcal{D}$ since $y(\mathbf{x}) \triangleq f(\mathbf{x}) + \epsilon$ and the fifth equality follows from the definition of KL divergence.
%
\section{Derivation of \eqref{elbog}}\label{a.elbog}
From \eqref{elbo}, 
%
\begin{equation*}
\begin{array}{l}
\mathcal{L}(q) \vspace{1mm}\\
 \displaystyle = \mathbb{E}_{q(\mathbf{f}_\mathcal{D}, \boldsymbol{\alpha}, k, \mathbf{g})}\left[\log p(\mathbf{y}_\mathcal{D} | \mathbf{f}_\mathcal{D})\right] 
\vspace{1mm}\\
\quad\displaystyle -\ \text{KL}\left[q(\mathbf{f}_\mathcal{D}, \boldsymbol{\alpha}, k, \mathbf{g})\|p(\mathbf{f}_\mathcal{D}, \boldsymbol{\alpha}, k, \mathbf{g})\right]
\vspace{1mm}\\
\displaystyle=\mathbb{E}_{p(k|\mathbf{g})q(\mathbf{g})} \left[\mathbb{E}_{p(\mathbf{f}_\mathcal{D} | \boldsymbol{\alpha}, k) q(\boldsymbol{\alpha})} [\log p(\mathbf{y}_\mathcal{D} | \mathbf{f}_\mathcal{D})] \right ]  \vspace{1mm}\\
\quad \displaystyle -\ \sum_{i=1}^K \text{KL} \left[q(\boldsymbol{\alpha}_i)||p(\boldsymbol{\alpha}_i)\right] - \text{KL} \left[q(\mathbf{g})||p(\mathbf{g})\right] \vspace{1mm}\\
=\displaystyle   \mathbb{E}_{q(\mathbf{g})} \left[\sum_{i=1}^{K}p(k_i|\mathbf{g})\ \mathbb{E}_{p(\mathbf{f}_\mathcal{D} | \boldsymbol{\alpha}, k_i) q(\boldsymbol{\alpha})} [\log p(\mathbf{y}_\mathcal{D} | \mathbf{f}_\mathcal{D})] \right]  \vspace{1mm}\\
\quad\displaystyle -\ \sum_{i=1}^K \text{KL} \left[q(\boldsymbol{\alpha}_i)||p(\boldsymbol{\alpha}_i)\right] - \text{KL} \left[q(\mathbf{g})||p(\mathbf{g})\right] \vspace{1mm}\\
\displaystyle =\mathbb{E}_{q(\mathbf{g})} \left[\sum_{i=1}^{K}p(k_i|\mathbf{g})\ \mathbb{E}_{p(\mathbf{f}_\mathcal{D} | \boldsymbol{\alpha}_i, k_i) q(\boldsymbol{\alpha}_i)} [\log p(\mathbf{y}_\mathcal{D} | \mathbf{f}_\mathcal{D})] \right] \vspace{1mm}\\
\quad \displaystyle -\ \mathbb{E}_{q(\mathbf{g})} \left[ \sum_{i=1}^K \text{KL} \left[q(\boldsymbol{\alpha}_i)||p(\boldsymbol{\alpha}_i)\right] \right] - \text{KL} \left[q(\mathbf{g})||p(\mathbf{g})\right] \vspace{1mm}\\
= \displaystyle \mathbb{E}_{q(\mathbf{g})}\left[\sum_{i=1}^{K} p(k_i | \mathbf{g})\ \mathcal{L}_{i}(q) \right]-\text{KL}\left[q(\mathbf{g})\|p(\mathbf{g})\right] 
\end{array}
\end{equation*}
where the second equality is due to~\eqref{p},~\eqref{q1}, and the additive property of KL divergence for independent distributions, and the fourth equality follows from the fact that $\mathbf{f}_\mathcal{D}$ is conditionally independent of $\boldsymbol{\alpha}_j$ for $j \neq i$ given kernel $k_i$ and $\boldsymbol{\alpha}_i$.
\section{Derivation of \eqref{grad}}\label{a.Li}
\subsection{Derivation of $\mathbb{E}_{ p(\mathbf{f}_\mathcal{D} | \boldsymbol{\alpha}_i, k_i) } [\log p(\mathbf{y}_\mathcal{D} | \mathbf{f}_\mathcal{D})]$}
Firstly, $p(\mathbf{f}_\mathcal{D} | \boldsymbol{\alpha}_i, k_i) =p(\mathbf{f}_\mathcal{D}|\mathbf{u}_i, \boldsymbol{\theta}_i, k_i)$ is a Gaussian with the following mean and covariance:
$$
\hspace{-1.7mm}
\begin{array}{rl}
\boldsymbol\mu_{\mathcal{D} | \boldsymbol{\alpha}_i} \hspace{-2.4mm}&\triangleq \Sigma^{\boldsymbol{\theta}_i}_{\langle\mathcal{D}, i\rangle \langle\mathcal{U}, i\rangle} (\Sigma^{\boldsymbol{\theta}_i}_{\langle\mathcal{U}, i\rangle \langle\mathcal{U}, i\rangle})^{-1} \mathbf{u}_i \vspace{1mm} \\
\Sigma_{\mathcal{D} | \boldsymbol{\alpha}_i} \hspace{-2.4mm}&\triangleq \Sigma^{\boldsymbol{\theta}_i}_{\langle\mathcal{D}, i\rangle \langle\mathcal{D}, i\rangle} - \Sigma^{\boldsymbol{\theta}_i}_{\langle\mathcal{D}, i\rangle \langle\mathcal{U}, i\rangle} (\Sigma^{\boldsymbol{\theta}_i}_{\langle\mathcal{U}, i\rangle \langle\mathcal{U}, i\rangle})^{-1} \Sigma^{\boldsymbol{\theta}_i}_{\langle\mathcal{U}, i\rangle\langle\mathcal{D}, i\rangle }
\end{array}
$$
where $\Sigma^{\boldsymbol{\theta}_i}_{\langle\mathcal{D}, i\rangle \langle\mathcal{U}, i\rangle} \triangleq (k^{\boldsymbol{\theta}_i}_i(\mathbf{x}, \mathbf{x}^\prime))_{\mathbf{x} \in \mathcal{X},\mathbf{x}^\prime \in \mathcal{U}}$, $\Sigma^{\boldsymbol{\theta}_i}_{\langle\mathcal{U}, i\rangle \langle\mathcal{U}, i\rangle} \triangleq (k^{\boldsymbol{\theta}_i}_i(\mathbf{x}, \mathbf{x}'))_{\mathbf{x}, \mathbf{x}' \in \mathcal{U}}$, and $k^{\boldsymbol{\theta}_i}_i(\mathbf{x}, \mathbf{x}')$ denotes $k_i(\mathbf{x}, \mathbf{x}')$ computed using hyperparameters $\boldsymbol{\theta}_i$. Then, 
\begin{equation*}
\begin{array}{l}
\mathbb{E}_{ p(\mathbf{f}_\mathcal{D} | \boldsymbol{\alpha}_i, k_i) } [\log p(\mathbf{y}_\mathcal{D} | \mathbf{f}_\mathcal{D})]
\vspace{1mm}\\
=\displaystyle\int p(\mathbf{f}_\mathcal{D} | \boldsymbol{\alpha}_i, k_i)\log \mathcal{N}(\mathbf{y}_{\mathcal{D}}|\mathbf{f}_{\mathcal{D}}, \sigma^2_n I)\ \text{d} \mathbf{f}_{\mathcal{D}}\vspace{1mm}\\
=\displaystyle\int p(\mathbf{f}_\mathcal{D} | \boldsymbol{\alpha}_i, k_i)\Bigg(-\frac{|\mathcal{D}|}{2}\log(2\pi\sigma^2_n)\vspace{1mm}\\
\quad \displaystyle -\frac{1}{2\sigma^2_n}\left(\mathbf{y}^\top_{\mathcal{D}}\mathbf{y}_{\mathcal{D}}-
2\mathbf{f}^\top_{\mathcal{D}}\mathbf{y}_{\mathcal{D}}+\mathbf{f}_{\mathcal{D}}^\top\mathbf{f}_{\mathcal{D}}\right)\Bigg)\ \text{d}\mathbf{f}_{\mathcal{D}}\vspace{1mm}\\
= \displaystyle -\frac{|\mathcal{D}|}{2}\log(2\pi\sigma^2_n)-\frac{1}{2\sigma^2_n}
(\mathbf{y}_{\mathcal{D}}^\top\mathbf{y}_{\mathcal{D}})\\
\quad \displaystyle +\frac{1}{\sigma^2_n} \mathbb{E}_{ p(\mathbf{f}_\mathcal{D} | \boldsymbol{\alpha}_i, k_i) } [\mathbf{y}^\top_{\mathcal{D}}\mathbf{f}_{\mathcal{D}}] - \frac{1}{2\sigma^2_n} \mathbb{E}_{ p(\mathbf{f}_\mathcal{D} | \boldsymbol{\alpha}_i, k_i) } [\mathbf{f}^\top_{\mathcal{D}}\mathbf{f}_{\mathcal{D}}] \vspace{1mm}\\
= \displaystyle -\frac{|\mathcal{D}|}{2}\log(2\pi\sigma^2_n)-\frac{1}{2\sigma^2_n}
(\mathbf{y}_{\mathcal{D}}^\top\mathbf{y}_{\mathcal{D}})\vspace{1mm}\\
\quad \displaystyle +\frac{1}{2\sigma^2_n}\left(2\mathbf{y}_{\mathcal{D}}^\top \boldsymbol\mu_{\mathcal{D} | \boldsymbol{\alpha}_i} -\boldsymbol\mu_{\mathcal{D}| \boldsymbol{\alpha}_i}^\top\boldsymbol\mu_{\mathcal{D} | \boldsymbol{\alpha}_i}
-\mathrm{tr}(\Sigma_{\mathcal{D} | \boldsymbol{\alpha}_i} )\right).
\end{array}
\end{equation*}
The gradient $\nabla_{\Phi_{i}}
\mathbb{E}_{ p(\mathbf{f}_\mathcal{D} | \boldsymbol{\alpha}_i, k_i) } [\log p(\mathbf{y}_\mathcal{D} | \mathbf{f}_\mathcal{D})]$ can thus be computed automatically using TensorFlow based on the above analytic equation.
\subsection{Derivation of $\text{KL}[q(\boldsymbol{\alpha}_i)||p(\boldsymbol{\alpha}_i)]$ }
Using the definition of KL divergence,
$$
\hspace{-1.7mm}
\begin{array}{l}
\displaystyle\text{KL}[q(\boldsymbol{\alpha}_i)||p(\boldsymbol{\alpha}_i)]\vspace{1mm}\\
\displaystyle =\text{KL}[q(\mathbf{u}_i)q(\boldsymbol{\theta}_i)||p(\mathbf{u}_i|\boldsymbol{\theta}_i)p(\boldsymbol{\theta}_i)]\vspace{1mm}\\
\displaystyle =\mathbb{E}_{q(\mathbf{u}_i)q(\boldsymbol{\theta}_i)}\left[\log \frac{q(\mathbf{u}_{i})q(\boldsymbol{\theta}_{i})}{p(\mathbf{u}_{i}|\boldsymbol{\theta}_{i})p(\boldsymbol{\theta}_{i})}\right]\vspace{1mm}\\
\displaystyle =\mathbb{E}_{q(\mathbf{u}_i)q(\boldsymbol{\theta}_i)}\left[\log q(\mathbf{u}_i)+\log \frac{q(\boldsymbol{\theta}_i)}{p (\boldsymbol{\theta}_i)}-\log p(\mathbf{u}_i|\boldsymbol{\theta}_i)\right]\vspace{1mm}\\
\displaystyle =-\mathbb{H}[q(\mathbf{u}_i)]+\text{KL}[q(\boldsymbol{\theta}_i)||p(\boldsymbol{\theta}_i)] -\mathbb{E}_{q(\mathbf{u}_i)q(\boldsymbol{\theta}_i)}[\log p(\mathbf{u}_i|\boldsymbol{\theta}_i)]
\end{array}
$$
where the entropy and the KL divergence terms can be computed analytically such that their gradients w.r.t.~$\Phi_i$ can be computed automatically using TensorFlow.
Though the last term cannot be computed analytically, we can approximate its gradient w.r.t.~$\Phi_i$ via the reparameterization trick (Section~\ref{opti}) by drawing i.i.d.~samples $\boldsymbol{\eta}^{(s)}_i$ for $s=1,\ldots,S$ from $q(\boldsymbol{\eta}_i)$:
$
\nabla_{\Phi_{i}} \mathbb{E}_{q(\mathbf{u}_i)q(\boldsymbol{\theta}_i)}[\log p(\mathbf{u}_i|\boldsymbol{\theta}_i)] = \nabla_{\Phi_{i}} \mathbb{E}_{q(\boldsymbol{\eta}_i)}[\log p(\mathbf{u}_i|\boldsymbol{\theta}_i)] =  \mathbb{E}_{q(\boldsymbol{\eta}_i)}[\nabla_{\Phi_{i}}\log p(\mathbf{u}_i|\boldsymbol{\theta}_i)]
\approx (1/S)\sum_{s=1}^{S} \nabla_{\Phi_{i}} \log p(\mathbf{u}_i^{(s)}|\boldsymbol{\theta}_i^{(s)})
$.
\section{Derivation of the Approximated Predictive Belief}\label{a.pred}
\subsection{Derivation of \eqref{predq}}
From~\eqref{predk},
$$
\hspace{-1.7mm}
\begin{array}{l}
 p(f(\mathbf{x}_*)|\mathbf{y}_\mathcal{D}) \vspace{1mm}\\
= \displaystyle  \sum_{i=1}^K \int p(f(\mathbf{x}_*) | \mathbf{f}_\mathcal{D}, \boldsymbol{\alpha}, k_i, \mathbf{g}) \ p( \mathbf{f}_\mathcal{D}, \boldsymbol{\alpha}, k_i, \mathbf{g} | \mathbf{y}_\mathcal{D}) \ \text{d} \mathbf{f}_\mathcal{D} \ \text{d} \boldsymbol{\alpha} \ \text{d} \mathbf{g} \vspace{1mm}\\
\approx \displaystyle \sum_{i=1}^K \int p(f(\mathbf{x}_*) | \mathbf{f}_\mathcal{D}, \boldsymbol{\alpha}, k_i, \mathbf{g}) \ 
p(\mathbf{f}_\mathcal{D} | \boldsymbol{\alpha}, k_i) \\
\qquad\quad\ \ \ \displaystyle p(k_i|\mathbf{g})\ q^*(\mathbf{g}) \prod_{j=1}^K q^*(\boldsymbol{\alpha}_j) \ \text{d} \mathbf{f}_\mathcal{D} \ \text{d} \boldsymbol{\alpha} \ \text{d} \mathbf{g} \vspace{1mm}\\
= \displaystyle \sum_{i=1}^K \int p(f(\mathbf{x}_*) | \boldsymbol{\alpha}_i, k_i) \ 
\left(\int p(\mathbf{f}_\mathcal{D} | \boldsymbol{\alpha}, k_i)\ \text{d} \mathbf{f}_\mathcal{D} \right) \vspace{1mm}\\
\qquad \displaystyle \left( \int p(k_i|\mathbf{g})\ q^*(\mathbf{g}) \ \text{d} \mathbf{g} \right) \left(\prod_{j \neq i} \int q^*(\boldsymbol{\alpha}_j) \ \text{d} \boldsymbol{\alpha}_j \right)  q^*(\boldsymbol{\alpha}_i)\ \text{d} \boldsymbol{\alpha}_i\vspace{1mm}\\
= \displaystyle \sum_{i=1}^K q^*(k_i) \int p(f(\mathbf{x}_*) | \boldsymbol{\alpha}_i, k_i)\  q^*(\boldsymbol{\alpha}_i)\ \text{d} \boldsymbol{\alpha}_i 
\end{array}
$$
where the approximation is due to~\eqref{q1},
the second equality follows from the DTC assumption that (a) $f(\mathbf{x}_*)$ is conditionally independent of $\mathbf{f}_\mathcal{D}$ given the inducing variables $\mathbf{u}_i$, and the assumption that (b) $f(\mathbf{x}_*)$ is conditionally independent of $\mathbf{g}$ and $\boldsymbol{\alpha}_j$ for $j \neq i$ given kernel $k_i$ and $\boldsymbol{\alpha}_i$, and the last equality is due to~\eqref{pk}.
\subsection{Derivation of \eqref{mv}}
In this subsection, we will first introduce how to approximate the predictive mean and variance of each $\int p(f(\mathbf{x}_*)|\boldsymbol{\alpha}_i, k_i) \ q^*(\boldsymbol{\alpha}_i) \ \text{d}\boldsymbol{\alpha}_i$ for $i = 1, \ldots, K$.

Following the test conditional in~\cite{Quinonero2005}, $p(f(\mathbf{x}_*)|\mathbf{u}_i, \boldsymbol{\theta}_i, k_i)$ is a Gaussian with the following mean and variance:
$$
\hspace{-1.7mm}
\begin{array}{rl}
\mu_{\langle\mathbf{x}_*, i\rangle | \boldsymbol{\alpha}_i} \hspace{-2.4mm}&\triangleq \Sigma^{\boldsymbol{\theta}_i}_{\langle\mathbf{x}_*, i\rangle \langle\mathcal{U}, i\rangle} (\Sigma^{\boldsymbol{\theta}_i}_{\langle\mathcal{U}, i\rangle \langle\mathcal{U}, i\rangle})^{-1} \mathbf{u}_i \vspace{1mm} \\
\sigma^2_{\langle\mathbf{x}_*, i\rangle | \boldsymbol{\alpha}_i } \hspace{-2.4mm}&\triangleq k^{\boldsymbol{\theta}_i}_i(\mathbf{x}_*, \mathbf{x}_*) - \Sigma^{\boldsymbol{\theta}_i}_{\langle\mathbf{x}_*, i\rangle \langle\mathcal{U}, i\rangle} (\Sigma^{\boldsymbol{\theta}_i}_{\langle\mathcal{U}, i\rangle \langle\mathcal{U}, i\rangle})^{-1} \Sigma^{\boldsymbol{\theta}_i}_{\langle\mathcal{U}, i\rangle\langle\mathbf{x}_*, i\rangle}
\end{array}
$$
where $\Sigma^{\boldsymbol{\theta}_i}_{\langle\mathbf{x}_*, i\rangle \langle\mathcal{U}, i\rangle} \triangleq (k^{\boldsymbol{\theta}_i}_i(\mathbf{x}_*, \mathbf{x}))_{\mathbf{x} \in \mathcal{U}}$, $\Sigma^{\boldsymbol{\theta}_i}_{\langle\mathcal{U}, i\rangle \langle\mathcal{U}, i\rangle} \triangleq (k^{\boldsymbol{\theta}_i}_i(\mathbf{x}, \mathbf{x}'))_{\mathbf{x}, \mathbf{x}' \in \mathcal{U}}$, $\Sigma^{\boldsymbol{\theta}_i}_{\langle\mathcal{U}, i\rangle\langle\mathbf{x}_*, i \rangle }\triangleq (\Sigma^{\boldsymbol{\theta}_i}_{\langle\mathbf{x}_*, i\rangle \langle\mathcal{U}, i\rangle})^\top$, and $k^{\boldsymbol{\theta}_i}_i(\mathbf{x}, \mathbf{x}')$ denotes $k_i(\mathbf{x}, \mathbf{x}')$ computed using hyperparameters $\boldsymbol{\theta}_i$.
Then,
\begin{equation}\label{pa}
\begin{array}{l}
\displaystyle\int p(f(\mathbf{x}_*)|\boldsymbol{\alpha}_i, k_i) \ q^*(\boldsymbol{\alpha}_i) \ \text{d}\boldsymbol{\alpha}_i  \vspace{1mm} \\
\displaystyle= \int \left(\int p(f(\mathbf{x}_*)|\mathbf{u}_i, \boldsymbol{\theta}_i, k_i) \ q^*(\mathbf{u}_i) \ \text{d}\mathbf{u}_i \right) q^*(\boldsymbol{\theta}_i)  \ \text{d}\boldsymbol{\theta}_i
\end{array}
\end{equation}
where the inner integration $\int p(f(\mathbf{x}_*)|\mathbf{u}_i, \boldsymbol{\theta}_i, k_i) \ q^*(\mathbf{u}_i) \ \text{d}\mathbf{u}_i$ can be computed analytically as a Gaussian with the following mean and variance:
$$
\hspace{-1.7mm}
\begin{array}{l}
\mu_{\langle \mathbf{x}_*, i \rangle | \boldsymbol{\theta}_i} \triangleq \Sigma^{\boldsymbol{\theta}_i}_{\langle\mathbf{x}_*, i\rangle \langle\mathcal{U}, i\rangle} (\Sigma^{\boldsymbol{\theta}_i}_{\langle\mathcal{U}, i\rangle \langle\mathcal{U}, i\rangle})^{-1} \mathbf{m}^*_{\mathbf{u}_i} \vspace{1mm}\\
\sigma^2_{\langle \mathbf{x}_*, i \rangle | \boldsymbol{\theta}_i} \triangleq k_i(\mathbf{x}_*, \mathbf{x}_*) - \Sigma^{\boldsymbol{\theta}_i}_{\langle\mathbf{x}_*, i\rangle \langle\mathcal{U}, i\rangle} (\Sigma^{\boldsymbol{\theta}_i}_{\langle\mathcal{U}, i\rangle \langle\mathcal{U}, i\rangle})^{-1} \Sigma^{\boldsymbol{\theta}_i}_{\langle\mathcal{U}, i\rangle\langle\mathbf{x}_*, i\rangle } \vspace{1mm} \\
+ \Sigma^{\boldsymbol{\theta}_i}_{\langle\mathbf{x}_*, i\rangle \langle\mathcal{U}, i\rangle} (\Sigma^{\boldsymbol{\theta}_i}_{\langle\mathcal{U}, i\rangle \langle\mathcal{U}, i\rangle})^{-1} \Sigma^*_{\mathbf{u}_i} (\Sigma^{\boldsymbol{\theta}_i}_{\langle\mathcal{U}, i\rangle \langle\mathcal{U}, i\rangle})^{-1} \Sigma^{\boldsymbol{\theta}_i}_{\langle\mathcal{U}, i\rangle\langle\mathbf{x}_*, i\rangle}\ .
\end{array}
$$
Though the outer integration w.r.t.~$\boldsymbol{\theta}_i$ cannot be computed analytically for all kernel types (except for some simple kernels such as SE and LIN), it can be approximated by drawing i.i.d.~samples $\boldsymbol{\theta}^{(1)}_i, \ldots, \boldsymbol{\theta}^{(S)}_i$ from $q^*(\boldsymbol{\theta}_i)$, which yields the following predictive mean and variance for \eqref{pa}:
$$
\begin{array}{rl}
\displaystyle \mu_{\langle \mathbf{x}_*, i \rangle} \hspace{-2.4mm}&\displaystyle\approx \frac{1}{S}\sum_{s=1}^S \mu_{\langle \mathbf{x}_*, i \rangle | \boldsymbol{\theta}^{(s)}_i} \vspace{1mm} \\
\displaystyle \sigma^2_{\langle \mathbf{x}_*, i \rangle} \hspace{-2.4mm}&\displaystyle\approx \frac{1}{S}\sum_{s=1}^S (\sigma^2_{\langle \mathbf{x}_*, i \rangle | \boldsymbol{\theta}^{(s)}_i} + \mu^2_{\langle \mathbf{x}_*, i \rangle | \boldsymbol{\theta}^{(s)}_i}) - \mu^2_{\langle \mathbf{x}_*, i \rangle} \ .
\end{array}
$$
The above results can be derived from $\mu_{\langle \mathbf{x}_*, i \rangle} = \mathbb{E}_{q^*(\boldsymbol{\theta}_i)}[\mathbb{E}_{p(f(\mathbf{x}_*)|\boldsymbol{\theta}_i, k_i)}[f(\mathbf{x}_*)]] = \mathbb{E}_{q^*(\boldsymbol{\theta}_i)}[\mu_{\langle \mathbf{x}_*, i \rangle | \boldsymbol{\theta}_i}]$ and
$\sigma^2_{\langle \mathbf{x}_*, i \rangle} = \mathbb{E}_{q^*(\boldsymbol{\theta}_i)}[\sigma^2_{\langle \mathbf{x}_*, i \rangle | \boldsymbol{\theta}_i} ] + \mathbb{V}_{q^*(\boldsymbol{\theta}_i)}[\mu_{\langle \mathbf{x}_*, i \rangle | \boldsymbol{\theta}_i}] = \mathbb{E}_{q^*(\boldsymbol{\theta}_i)}[\sigma^2_{\langle \mathbf{x}_*, i \rangle | \boldsymbol{\theta}_i} + \mu^2_{\langle \mathbf{x}_*, i \rangle | \boldsymbol{\theta}_i}] - \mu^2_{\langle \mathbf{x}_*, i \rangle}\ $. Then,~\eqref{mv} can be derived in the same manner by marginalizing out $k$ using~$q^*(k_i)$. 
\section{Details of Experimental Setup}
\subsection{Composite Kernels}\label{a.kernel}
The composite kernels in $\mathcal{K}_{12}$ (Section~\ref{syn}) are shown in Table \ref{tab.kernels} below:
\begin{table}[h]
	\centering
	\caption{Composite kernel types in $\mathcal{K}_{12}$.}
	\begin{tabular}{c|c}
		\hline
		   & Composite Kernel \\
		\hline
		 k1 & $\text{LIN}+\text{RQ}$ \\
		 k2 & $\text{LIN}\times\text{RQ}+\text{LIN}$ \\
		 k3 & $\text{LIN}\times\text{RQ}+\text{PER}$\\
		 k4 & $\text{PER}+\text{RQ}+\text{SE}$\\
		 k5 & $\text{PER}+\text{LIN}+\text{RQ}$\\
		 k6 & $\text{PER}+\text{PER}+\text{SE}$\\
		 k7 & $\text{PER}\times\text{SE}+\text{SE}$\\
		 k8 & $\text{PER}\times\text{RQ}+\text{SE}$\\
		 k9 & $\text{PER}\times\text{LIN}+\text{SE}$\\
		 k10 & $\text{PER}\times\text{LIN}\times\text{SE}$\\
		 k11 & $\text{PER} \times \text{LIN} \times \text{RQ}$\\
		 k12 & $(\text{PER} + \text{RQ})\times \text{LIN} $\\
		\hline
	\end{tabular}
	\label{tab.kernels}
\end{table}


The following true kernels are used to generate the synthetic datasets.
The three base kernels in $(\text{PER} + \text{RQ})\times \text{LIN} $ are
$$
\begin{array}{rl}
k_{\text{PER}}(\mathbf{x},\mathbf{x}^{\prime})\hspace{-2.4mm}&\displaystyle \triangleq 0.1^{2}\exp\left(-\frac{2\sin^{2}(\pi|\mathbf{x}-\mathbf{x}^{\prime}|/2\pi)}{{2}^{2}}\right)\\
k_{\text{RQ}}(\mathbf{x},\mathbf{x}^{\prime})\hspace{-2.4mm}&\displaystyle \triangleq 0.1^{2}\left(1+\frac{(\mathbf{x}-\mathbf{x}^{\prime})^2}{2\times1\times
	{3}^{2}}\right)^{-1}\\
k_{\text{LIN}}(\mathbf{x},\mathbf{x}^{\prime})\hspace{-2.4mm}&\displaystyle \triangleq \frac{\mathbf{x}\mathbf{x}^{\prime}}{{5}^{2}}\ .
\end{array}
$$
The three base kernels in $\text{PER} \times \text{LIN} \times \text{RQ}$ are
$$
\begin{array}{rl}
k_{\text{PER}}(\mathbf{x},\mathbf{x}^{\prime})\hspace{-2.4mm}&\displaystyle \triangleq 0.1^{2}\exp\left(-\frac{2\sin^{2}(\pi|\mathbf{x}-\mathbf{x}^{\prime}|/2\pi)}{{1}^{2}}\right)\\
k_{\text{RQ}}(\mathbf{x},\mathbf{x}^{\prime})\hspace{-2.4mm}&\displaystyle \triangleq 0.1^{2}\left(1+\frac{(\mathbf{x}-\mathbf{x}^{\prime})^2}{2\times1\times
	{8}^{2}}\right)^{-1}\\
k_{\text{LIN}}(\mathbf{x},\mathbf{x}^{\prime})\hspace{-2.4mm}&\displaystyle \triangleq \frac{\mathbf{x}\mathbf{x}^{\prime}}{{3}^{2}}\ .
\end{array}
$$

\subsection{Additional Experimental Results}\label{a.result}
Tables~\ref{tab.swiss} and~\ref{tab.IEQ} show the final RMSEs incurred by VBKS and VBKS-s for $10$ independent runs that correspond to the results in Figs.~\ref{fig.swiss}b and~\ref{fig.IEQ}b (i.e., showing averaged RMSEs with standard deviation), respectively. 
For both datasets, it can be observed that VBKS consistently incurs smaller RMSEs than VBKS-s in all $10$ independent runs.
\begin{table}[h]
\centering
\caption{RMSEs (MWh) for $10$ runs for Swissgrid dataset.}
\begin{tabular}{c|c|c}
\hline
run & VBKS-s & VBKS  \\ \hline
1   & 160.6  & 155.9 \\
2   & 162.1  & 159.3 \\
3   & 162.0  & 159.9 \\
4   & 159.4  & 156.2 \\
5   & 159.0  & 156.2 \\
6   & 159.2  & 156.7 \\
7   & 159.2  & 156.9 \\
8   & 163.3  & 158.4 \\
9   & 158.4  & 155.0 \\
10  & 158.6  & 155.0 \\
\hline
\end{tabular}
\label{tab.swiss}
\end{table}

\begin{table}[h]
\centering
\caption{RMSEs ($^\circ$C) for $10$ runs for IEQ dataset.}
\begin{tabular}{c|c|c}
\hline
run & VBKS-s & VBKS  \\ \hline
1   & $0.416$ & 0.380 \\
2   & 0.380  & 0.371 \\
3   & 0.383  & 0.373 \\
4   & 0.386  & 0.374 \\
5   & 0.379  & 0.367 \\
6   & 0.386  & 0.372 \\
7   & 0.380  & 0.370 \\
8   & 0.384  & 0.371 \\
9   & 0.395  & 0.378 \\
10  & 0.390  & 0.377 \\ \hline
\end{tabular}
\label{tab.IEQ}
\end{table}

\fi

\end{document}